\documentclass{article}

\PassOptionsToPackage{numbers,compress}{natbib}
\PassOptionsToPackage{table}{xcolor}
\usepackage[preprint]{neurips_2026}

\usepackage[utf8]{inputenc}
\usepackage[T1]{fontenc}
\usepackage{hyperref}
\usepackage{url}
\usepackage{booktabs}
\usepackage{graphicx}
\graphicspath{{./}{figures_new/}}
\usepackage{amsmath}
\usepackage{amsfonts}
\usepackage{amssymb}
\usepackage{nicefrac}
\usepackage{microtype}
\usepackage{array}
\usepackage{multirow}
\usepackage{tabularx}
\usepackage{xcolor}
\usepackage{float}
\usepackage{placeins}
\usepackage{algorithm}
\usepackage{algpseudocode}
\usepackage{pifont}
\usepackage{xspace}
\newcommand{\modelname}{\textsc{Sira}\xspace}
\newcommand{\cmark}{\ding{51}}
\newcommand{\xmark}{\ding{55}}
\newcolumntype{Y}{>{\centering\arraybackslash}X}

\title{Do We Really Need External Tools to Mitigate Hallucinations? \modelname: Shared-Prefix Internal Reconstruction of Attribution}

\author{%
  Tian Qin$^{1,2*}$,
  Junzhe Chen$^{1*}$,
  Yuqing Shi$^{3}$,
  Tianshu Zhang$^{1}$,
  Qiang Ju$^{4}$,
  Lijie Wen$^{1\dagger}$\\
  $^{1}$Tsinghua University,
  $^{2}$The University of Sydney,
  $^{3}$Stanford University,
  $^{4}$Baichuan AI.\\
  $^{*}$Equal contribution.\quad $^{\dagger}$Corresponding author.\\
  \texttt{tqin0515@uni.sydney.edu.au}\\
  \texttt{chenjz24@mails.tsinghua.edu.cn}\\
  \texttt{wenlj@tsinghua.edu.cn}\\
}

\begin{document}
\raggedbottom
\maketitle

\begin{abstract}
Large vision-language models (LVLMs) often hallucinate when language priors dominate weak or ambiguous visual evidence. Existing contrastive decoding methods mitigate this problem by comparing predictions from the original image with those from externally perturbed visual inputs, but such references can introduce off-manifold artifacts and require costly extra forward passes. We propose SIRA, a training-free internal contrastive decoding framework that constructs a counterfactual reference inside the same LVLM by exploiting the staged information flow of multimodal transformers. Instead of removing visual information from the input, SIRA first lets image and text tokens interact through a shared prefix, forming an aligned multimodal state that preserves prompt interpretation, decoding history, positional structure, and early visual grounding. It then forks a counterfactual branch in later transformer layers, where attention to image-token positions is masked. This branch retains the shared multimodal context but lacks continued access to fine-grained visual evidence, yielding a language-prior-dominated internal reference for token-level contrast. During decoding, SIRA suppresses tokens that remain strong without late visual access and favors predictions whose advantage depends on the full visual pathway. Experiments on POPE, CHAIR, and AMBER with Qwen2.5-VL and LLaVA-v1.5 show that SIRA consistently reduces hallucinations while preserving descriptive coverage and incurring lower overhead than two-pass contrastive decoding. SIRA requires no training, external verifier, or perturbed input, and applies to open-weight LVLMs with white-box inference access.
\end{abstract}

\section{Introduction}
\label{sec:intro}
Large vision-language models (LVLMs) have made rapid progress on multimodal understanding and generation tasks~\citep{li2023blip2, geminiteam2025gemini, liu2024improvedbaselinesvisualinstruction, Ye_2024, bai2025qwen25vltechnicalreport}. Despite this progress, they remain vulnerable to hallucination~\citep{li2023evaluating, rohrbach2018object, Tong_2024, Lee_2025}: the model may describe objects, attributes, or relations that are plausible from linguistic context but unsupported by the image. Such errors often occur when visual evidence is weak or ambiguous, allowing high-probability language continuations to dominate token prediction. Hallucination mitigation therefore requires not only stronger visual representations, but also better control over how visual evidence and language priors are integrated inside the model during decoding.

Recent work addresses this problem through contrastive decoding~\citep{Li_2023, chuang2024dola, leng2024mitigating, wang-etal-2024-mitigating, Kim_2024, liu-etal-2025-multi, deng-yang-2025-maskcd, tong2025mitigatinghallucinationmultimodalllms, chen-etal-2026-mask, wan2025only}. These methods compare the model's prediction under the original multimodal input with a reference prediction obtained from a degraded, perturbed, or otherwise counterfactual condition, and then favor tokens whose advantage is attributed to visual evidence rather than generic language plausibility. This token-level view is appealing because hallucinations are ultimately realized through decoding decisions. However, many LVLM-specific contrastive methods construct their reference externally in the input space, for example by blurring, masking, or otherwise weakening the image~\citep{leng2024mitigating, wang-etal-2024-mitigating, deng-yang-2025-maskcd, chen-etal-2026-mask}. This design has two limitations. First, it often requires an additional full forward pass. Second, input-side perturbations may introduce off-manifold artifacts, so the resulting logit discrepancy may reflect perturbation-induced mismatch rather than clean visual attribution~\citep{leng2024mitigating, wang-etal-2024-mitigating, deng-yang-2025-maskcd, chen-etal-2026-mask, huo2025selfintrospective, yin2026the}.

\begin{figure}[t]
\centering
\includegraphics[width=.9\textwidth]{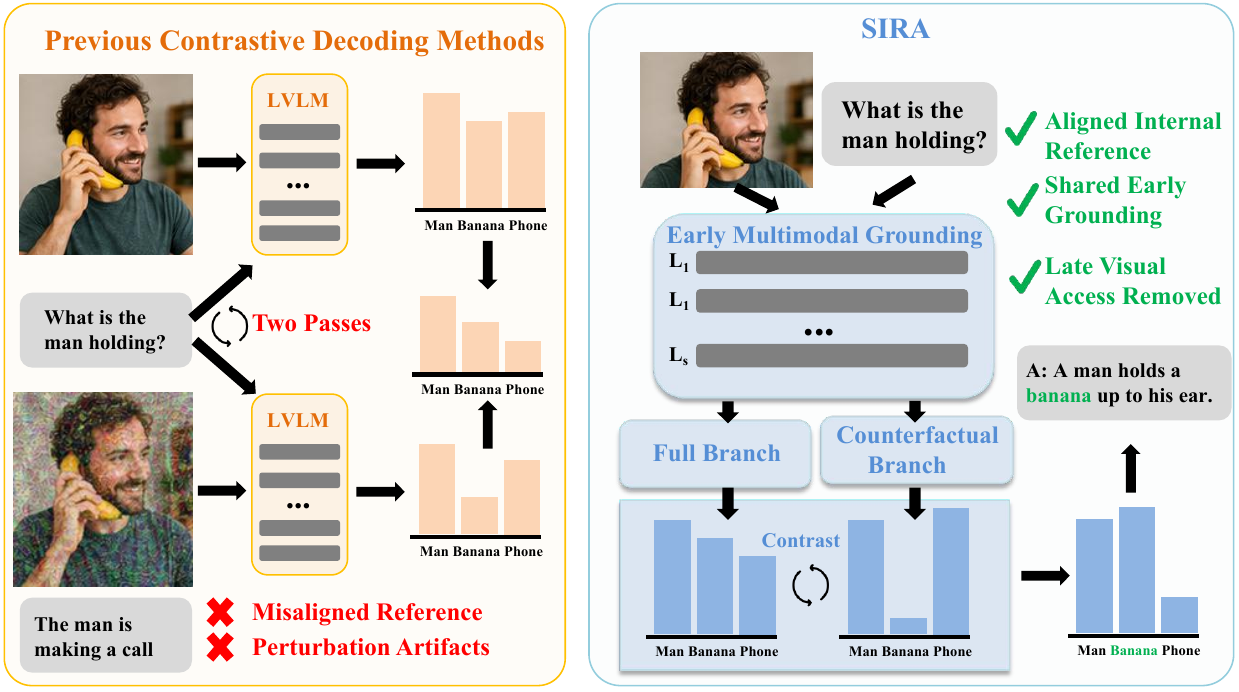}
\vspace{-3.5mm}
\caption{Comparison between representative prior inference-time mitigation methods and \modelname.}
\vspace{-6.5mm}
\label{fig:sira-intro}
\end{figure}

In this paper, we ask a different question: \emph{does contrastive decoding really need an externally constructed counterfactual input?} We argue that the key requirement is an aligned reference that reveals how the same multimodal state would continue when later visual access is restricted. This motivates us to construct the reference by manipulating information flow inside the LVLM. Multimodal transformers integrate image and text information progressively: early layers inject visual evidence into instruction and text-token states, while later layers further use image-token evidence to refine token-level decisions. Therefore, a useful counterfactual branch should not be separated from the input layer, which would destroy early visual-to-text grounding and make the two branches semantically misaligned. It should also not be separated too late, when visual evidence has already been absorbed into hidden states and masking image tokens has little effect.

We propose \textbf{S}hared-Prefix \textbf{I}nternal \textbf{R}econstruction of \textbf{A}ttribution (\modelname), a training-free internal contrastive decoding framework for LVLMs. As illustrated in Figure~\ref{fig:sira-intro}, \modelname constructs a synchronized counterfactual branch within the same inference trajectory instead of perturbing the image input. The full and counterfactual branches share the prompt, model parameters, decoding history, positional structure, and early-layer computation. After this shared prefix, the counterfactual branch is prevented from further attending to image-token positions. The shared prefix is therefore not merely an efficiency device: it preserves prompt interpretation and early multimodal grounding, so both branches start from the same image-conditioned semantic state. After branching, the main difference is whether later layers can continue to access fine-grained visual evidence. At each decoding step, \modelname contrasts the logits of the full branch with those of the counterfactual branch. Tokens that remain strong without post-boundary visual access are down-weighted, since they are more likely supported by language priors or generic textual continuations. Tokens whose advantage appears primarily in the full branch are favored as visually grounded. In this way, \modelname performs online internal prior cancellation between two continuations of the same early-grounded multimodal state, without additional training, auxiliary verifiers, external image perturbations, or a second full multimodal forward pass.

Our contributions are summarized as follows:

$\bullet$ We revisit contrastive decoding for LVLM hallucination mitigation from an information-flow perspective, arguing that an effective reference should preserve early multimodal grounding while restricting later image-token access.

$\bullet$ We propose \modelname, a training-free internal contrastive decoding framework that constructs a reference inside the same LVLM via shared-prefix late branching and post-boundary image-token masking.

$\bullet$ We provide comprehensive empirical evidence on POPE, CHAIR, and AMBER across Qwen2.5-VL and LLaVA-v1.5, showing that \modelname consistently reduces hallucination while preserving descriptive coverage and incurring lower overhead than two-pass contrastive decoding.

\section{Related Work}
\label{sec:related}

\subsection{Large Vision-Language Models}
\label{sec:related-lvlm}

Large vision-language models (LVLMs) extend large language models~\citep{touvron2023llama, touvron2023llama2, achiam2023gpt4} with visual encoders and feature projectors to handle multimodal tasks~\citep{li2023blip2, geminiteam2025gemini, liu2024improvedbaselinesvisualinstruction, Ye_2024, bai2025qwen25vltechnicalreport}. While such designs effectively inject visual information into the LLM backbone, how multimodal fusion actually unfolds across transformer layers has only recently been studied. \citet{yin2025lifting} show that shallow layers inject visual information into instruction tokens to form cross-modal semantic representations, while deeper layers intensify intra-visual interactions that aggregate residual features. \citet{kaduri2025whats} further demonstrate that cross-modal information transfer concentrates in the middle layers, with early and late layers contributing only marginally. Consistent with evidence that factual associations localize to specific transformer layers~\citep{meng2023locatingeditingfactualassociations}, these findings collectively suggest a \emph{staged fusion pattern}: early-layer visual--textual injection, middle-layer cross-modal consolidation, and a late-layer decision regime where visual evidence has largely been absorbed. Despite this progress, LVLMs still suffer from hallucinations~\citep{li2023evaluating, rohrbach2018object, Tong_2024, Lee_2025} that limit their real-world deployment.

\subsection{Hallucination Mitigation in LVLMs}
\label{sec:related-hallucination}

LVLMs produce cross-modal inconsistencies between visual inputs and responses, known as hallucinations~\citep{Huang_2025, li2023evaluating, Zhong_2024, Gunjal_2024}. Existing mitigation methods differ mainly in where they introduce the corrective signal. Training-based approaches fine-tune the backbone with hallucination-aware data, preference feedback, or auxiliary objectives~\citep{Yu_2024, Zhang_2024_2, Jiang_2024, Sun_2024, Yu_2024_2}, but they require additional supervision and model updates. Inference-time methods avoid retraining by manipulating attention, adding external modules, or applying trusted interventions~\citep{huang2024opera, An_2025, Woo_2025, Liu_2024_3, chen2025ict, Yin_2024, Favero_2024, Yue_2024}; these methods are easier to deploy than fine-tuning, but they can still add per-step control overhead or rely on components outside the base LVLM.

Contrastive and reference-based decoding methods target the token-selection step more directly. General contrastive decoding and DoLa contrast logits from different generators or layers~\citep{Li_2023, chuang2024dola}, while LVLM-specific variants build references from perturbed images, instruction changes, self-generated descriptions, frequency-filtered views, masked image heads, self-introspective signals, layer contrasts, object-aligned auxiliary views, or one-layer interventions~\citep{leng2024mitigating, wang-etal-2024-mitigating, Kim_2024, liu-etal-2025-multi, deng-yang-2025-maskcd, huo2025selfintrospective, tong2025mitigatinghallucinationmultimodalllms, chen-etal-2026-mask, wan2025only}. These methods share the useful idea that hallucination can be reduced by comparing the original prediction with a reference that exposes language-prior bias. However, the reference is often constructed outside the original inference path or from a condition that is not fully aligned with the original multimodal state, so the logit discrepancy can mix visual attribution with artifacts of perturbation, auxiliary views, or layer mismatch~\citep{yin2026the}. Motivated by the staged fusion pattern (Section~\ref{sec:related-lvlm}), we instead construct the reference \emph{inside} the same model: a shared prefix preserves prompt interpretation and early semantic grounding, and post-boundary image-token masking isolates the visual contribution in logit space. Because both branches share weights, prompt, decoding history, and early multimodal interpretation, \modelname avoids introducing a separate off-manifold input condition and yields a more direct token-level attribution proxy than externally constructed references.

\section{Method}
\label{sec:method}

\modelname augments a single LVLM with a counterfactual branch that shares early-layer computation with the standard multimodal branch and diverges only after a chosen split boundary. The shared prefix preserves semantic alignment: both branches inherit the same prompt interpretation, decoding history, positional structure, and early multimodal grounding. After the boundary, \modelname masks image-token access in the counterfactual branch, so the two branches differ mainly in whether visual evidence can continue to shape token prediction. At every decoding step, \modelname obtains this contrast inside the same model, without launching a second full forward pass over a perturbed image or relying on an external reference model. We describe the branch design (Section~\ref{sec:method-framework}), the masked counterfactual construction (Section~\ref{sec:method-cf}), and the contrastive decoding rule (Section~\ref{sec:method-decode}).

\subsection{Internal Branch Design}
\label{sec:method-framework}

\begin{figure*}[t]
\centering
\includegraphics[width=\textwidth]{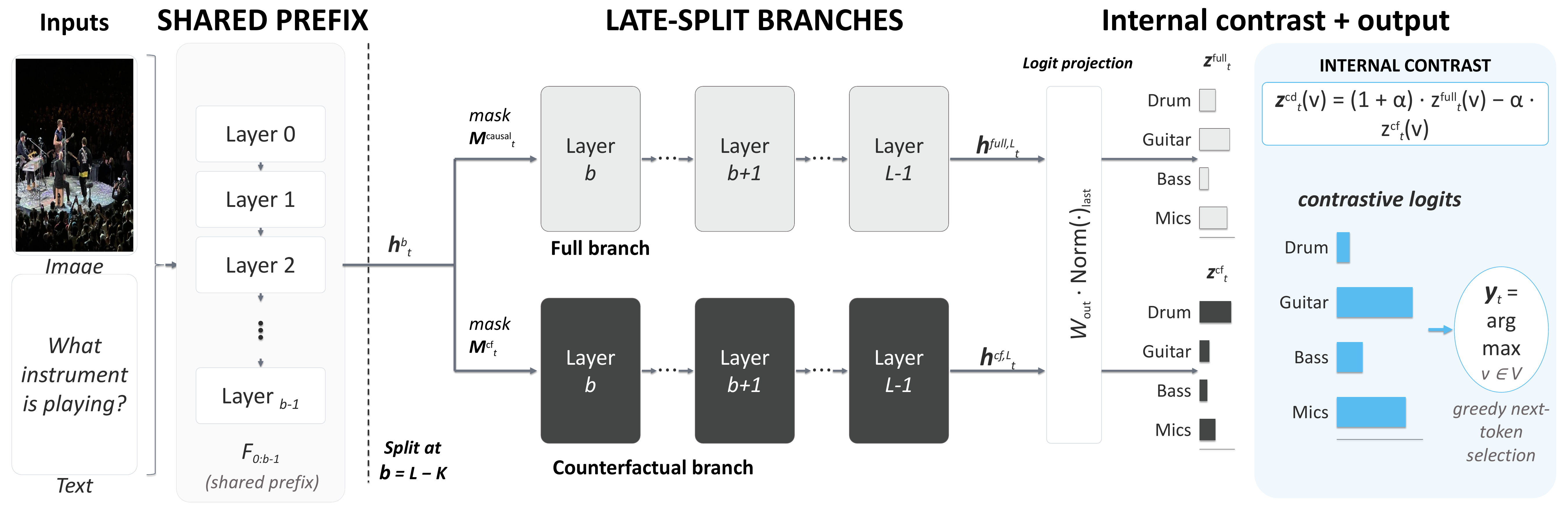}
\vspace{-5mm}

\caption{\modelname overview. $B$ in the figure denotes boundary $b{=}L{-}K$ in the text. Layer numbers are illustrative; $L$ is backbone-dependent.}
\vspace{-5mm}

\label{fig:sira-overview}
\end{figure*}

As shown in Figure~\ref{fig:sira-overview}, \modelname centers on a within-model comparison: each multimodal prediction is compared with an internal reference dominated by textual context and language priors. This avoids mismatches from external verifiers or auxiliary models while keeping inference lightweight. Specifically, \modelname maintains two synchronized decoding branches: a \textbf{full branch} that performs standard multimodal inference, and a \textbf{counterfactual branch} that uses the same prompt, parameters, and history but is blocked from attending to image-token positions from boundary $b$ onward. To reduce cost, layers $0$ to $b-1$ are computed once, and branch separation is applied only in the final $K$ layers. The discrepancy between the two branches thus provides an approximate attribution signal, measuring how much a candidate token relies on visual evidence rather than language priors.

Let $\mathbf{x}_{1:S}$ denote the unified prompt tokens after preprocessing, and let $\mathcal{P}_{\mathrm{img}}\subseteq\{1,\dots,S\}$ denote the positions occupied by image tokens. At decoding step $t$, the current sequence is
$\mathbf{u}_t=[\mathbf{x}_{1:S};y_{<t}]$.
The set $\mathcal{P}_{\mathrm{img}}$ remains fixed during decoding, since generated tokens are ordinary text tokens. We split the $L$-layer backbone at boundary $b = L-K$, where $K$ denotes the number of post-boundary layers that participate in branch separation. Let $F_{0:b-1}$ denote the shared prefix layers and $F_{b:L-1}$ denote the post-boundary layers. \modelname first obtains a shared boundary representation $\mathbf{h}^{b}_t = F_{0:b-1}(\mathbf{u}_t)$.
The two branches are then defined as two continuations of the same boundary state:
\begin{equation}
    \mathbf{h}^{\mathrm{full},L}_t
=F_{b:L-1}(\mathbf{h}^{b}_t;M^{\mathrm{causal}}_t),
\qquad
\mathbf{h}^{\mathrm{cf},L}_t
=F_{b:L-1}(\mathbf{h}^{b}_t;M^{\mathrm{cf}}_t).
\end{equation}
The boundary balances two requirements: a shared prefix deep enough to form a stable multimodal interpretation, and enough post-boundary layers to remove visual evidence before it is fully absorbed into text-token representations. Following the staged fusion pattern (Section~\ref{sec:related-lvlm}), we place $b$ just before the main consolidation stage; Section~\ref{sec:disc-split} confirms this empirically.

The shared prefix is what makes the contrast interpretable. Rather than comparing the multimodal model to an independently constructed text-only or corrupted-image pass, \modelname compares two continuations of the same internal state: both branches inherit identical prompt processing, decoding history, positional encoding, and early semantic grounding, and the only deliberate intervention after $b$ is whether direct image-token access remains available. The resulting logit discrepancy $\mathbf{z}^{\mathrm{full}}_t-\mathbf{z}^{\mathrm{cf}}_t$ therefore reflects the post-boundary visual contribution rather than unrelated representation mismatch; residual drift from masking is empirically small (Section~\ref{sec:disc-ref-quality}). The next subsection formalizes the post-boundary visual-access constraint with masked attention.

\subsection{Counterfactual Branch Construction}
\label{sec:method-cf}

The counterfactual branch must suppress further visual access while preserving the model's linguistic preference and textual reasoning path. \modelname does not alter the prompt, parameters, or output head; it only removes the post-boundary route through which image tokens can still influence the decoder, namely attention to image-token positions. This mask-level intervention keeps the branch on the same model dynamics as the full branch and avoids a second full forward pass over a corrupted input, yielding a cleaner counterfactual than input-space perturbations.

Let $q$ and $k$ index positions in the current sequence $\mathbf{u}_t$, and let $M^{\mathrm{causal}}_t(q,k)$ denote standard causal-and-padding validity. The counterfactual mask removes image-token access by excluding image positions both as keys and as query rows in the post-boundary branch, so that image-token states can neither be read by other queries nor produce queries that feed the text stream. Formally, we define counterfactual validity as
\begin{equation}
M^{\mathrm{cf}}_t(q,k)=
\begin{cases}
1, & \text{if } M^{\mathrm{causal}}_t(q,k)=1,\ q\notin\mathcal{P}_{\mathrm{img}},\ k\notin\mathcal{P}_{\mathrm{img}},\\
0, & \text{otherwise,}
\end{cases}
\end{equation}
where $\mathcal{P}_{\mathrm{img}}$ denotes the image-token positions in the original prompt; generated tokens are text positions. The validity mask is converted into the model's additive attention mask, where valid entries receive $0$ and invalid entries receive a large negative value. The mask therefore acts in two phases. In the post-boundary prefix pass over layers $b,\ldots,L-1$ on the original prompt, the $q\notin\mathcal{P}_{\mathrm{img}}$ condition is active and non-trivial: it removes image tokens as query rows, so image-token states are prevented from producing queries that read the prompt and from writing into the post-boundary KV cache. During per-step autoregressive decoding, every new query is a generated text token, so the $q\notin\mathcal{P}_{\mathrm{img}}$ condition is satisfied automatically and only the $k\notin\mathcal{P}_{\mathrm{img}}$ condition remains active, preventing new tokens from attending back to image-token keys. As a result, after boundary $b$, the counterfactual branch keeps the shared semantic prefix but cannot further use image-token positions, making it a language-prior-dominated internal reference.

Computationally, this design avoids an extra full forward pass by sharing the pre-boundary layers and branching only after a mid-layer split. If a standard decoding step costs $L$ layers, \modelname adds only the post-boundary $K$ layers for the counterfactual stream, giving an idealized per-token cost ratio of $(L+K)/L = 1 + K/L$; for a middle split this is roughly $1.5\times$ rather than the $\sim\!2\times$ cost of a second full-model pass. Appendix~\ref{app:efficiency} reports the corresponding latency measurements.

From layer $b$ onward, the two branches are evaluated in parallel under branch-specific masks. The computation is explicitly batched so that the model compares aligned states at every decoding step without changing the underlying backbone. We preserve positional consistency by duplicating the model's native positional encoding inputs across the two branches. Thus, the only intended difference between the two streams is visual accessibility after the split. With the two branches aligned in this way, decoding can directly compare their token-score distributions.

\subsection{Internal Contrastive Decoding}
\label{sec:method-decode}

With the two branches aligned, \modelname contrasts their logits at each step. At decoding step $t$, let $\mathbf{h}^{\mathrm{full},L}_t,\mathbf{h}^{\mathrm{cf},L}_t \in \mathbb{R}^{|\mathbf{u}_t|\times d}$ denote the final-layer hidden states of the full and counterfactual branches for the current sequence $\mathbf{u}_t$, and let $(\cdot)_{\mathrm{last}}$ denote the row at the last sequence position $|\mathbf{u}_t|$, i.e., the representation that drives next-token prediction. The two branch-wise logits are
\begin{equation}
\mathbf{z}^{\mathrm{full}}_t
=W_{\mathrm{out}}\operatorname{Norm}(\mathbf{h}^{\mathrm{full},L}_t)_{\mathrm{last}},
\qquad
\mathbf{z}^{\mathrm{cf}}_t
=W_{\mathrm{out}}\operatorname{Norm}(\mathbf{h}^{\mathrm{cf},L}_t)_{\mathrm{last}},
\end{equation}
where $\operatorname{Norm}(\cdot)$ and $W_{\mathrm{out}}$ denote the model's native final normalization and output projection, shared by both branches. \modelname compares these two predictions in logit space so that tokens supported by the image are amplified relative to those that remain strong under the counterfactual reference. We define the internal contrast $\Delta_t(v) = z^{\mathrm{full}}_t(v) - z^{\mathrm{cf}}_t(v)$ for vocabulary index $v$ and form the decoding logit
\begin{equation}
\label{alg:logit}
z^{\mathrm{cd}}_t(v) = z^{\mathrm{full}}_t(v) + \alpha\,\Delta_t(v)
= (1+\alpha)z^{\mathrm{full}}_t(v) - \alpha z^{\mathrm{cf}}_t(v),
\end{equation}
where $\alpha\ge 0$ controls contrast strength. When $\alpha=0$, the rule reduces to standard decoding; as $\alpha$ increases, tokens that remain strong in the counterfactual branch receive a larger penalty, while tokens whose advantage comes specifically from visual grounding are emphasized. This operation is best understood as approximate baseline subtraction in logit space: $\mathbf{z}^{\mathrm{cf}}_t$ serves as a language-prior-dominated reference constructed in Section~\ref{sec:method-cf}, and subtracting it from $\mathbf{z}^{\mathrm{full}}_t$ yields a proxy for the incremental contribution of visual evidence to each token score. The resulting contrast counteracts language-template bias in real time while preserving the original model parameters and input.

The next token is selected greedily from the contrastive logits, $y_t = \arg\max_{v\in\mathcal{V}} z^{\mathrm{cd}}_t(v)$.
The decoding loop maintains a standard full-branch cache alongside a late-layer branch cache for the stacked two-stream computation: after prefill, each step updates the full branch, then the late-layer parallel branches, and finally applies contrastive fusion to select the next token. Together, the internal branching (Section~\ref{sec:method-framework}), masked counterfactual construction (Section~\ref{sec:method-cf}), and decoding rule above form a training-free, token-level control mechanism that operates within a single model and avoids any second full forward pass.

\section{Experiments}
\label{sec:experiments}

\subsection{Experimental Setup}
\label{sec:exp-setup}

\noindent \textbf{Datasets and Metrics.}
We evaluate \modelname on three complementary hallucination benchmarks that together cover discriminative question answering, short captioning, and open-ended generation.

\textbf{POPE}~\citep{li2023evaluating} probes object-level hallucination through balanced Yes/No questions about the presence of specific objects in an image. Each split is 50\% positive and 50\% negative queries, and three negative-sampling strategies control how strongly textual co-occurrence biases the model toward the wrong answer. Following the standard POPE setup, queries are drawn from MSCOCO~\citep{lin2014microsoft}, A-OKVQA~\citep{schwenk2022okvqa}, and GQA~\citep{hudson2019gqa}, and we report Accuracy and F1 on each (dataset, split) pair.

\textbf{CHAIR}~\citep{rohrbach2018object} evaluates object hallucination in free-form captioning by checking whether the objects mentioned in a generated caption actually appear in the image. We report the sentence-level rate CHAIRs and the instance-level rate CHAIRi (lower is better) on a random subset of 500 images from the COCO 2014 validation set, at two generation lengths, Len64 and Len512.

\textbf{AMBER}~\citep{wang2023amber} is a multidimensional captioning benchmark; we use its generation task and report four metrics. CHAIR, Hal, and Cog capture different aspects of hallucinated content (lower is better), while Cover measures descriptive coverage of the image (higher is better). Reporting Hal and Cog jointly with Cover lets us check whether a method reduces hallucination by shortening or hedging the caption rather than by correcting the offending tokens.

\noindent \textbf{Baselines.} We adopt LLaVA-v1.5-7B~\citep{liu2024improvedbaselinesvisualinstruction} and Qwen2.5-VL-7B~\citep{bai2025qwen25vltechnicalreport} as our base LVLMs. We compare \modelname against inference-time methods that mitigate hallucinations by intervening at the decoding stage: DoLa~\citep{chuang2024dola}, OPERA~\citep{huang2024opera}, VCD~\citep{leng2024mitigating}, ICT~\citep{chen2025ict}, MaskCD~\citep{deng-yang-2025-maskcd}, and SID~\citep{huo2025selfintrospective}. Appendix~\ref{app:baseline-details} summarizes their venues, intervention mechanisms, and how they differ from \modelname. Implementation details can be seen in Appendix~\ref{app:imp}.

\subsection{POPE}
\label{sec:exp-pope}

\begin{table}[!tbp]
\centering
\scriptsize
\setlength{\tabcolsep}{1.2pt}
\renewcommand{\arraystretch}{1.22}
\resizebox{\textwidth}{!}{%
\begin{tabular}{ll@{\hspace{4pt}}l cccccccc|cccccccc}
\toprule
& & & \multicolumn{8}{c|}{Qwen2.5-VL} & \multicolumn{8}{c}{LLaVA-v1.5} \\
\cmidrule(lr){4-11} \cmidrule(lr){12-19}
Dataset & Setting & & Base & DoLa & OPERA & VCD & ICT & MaskCD & SID & \textbf{\modelname} & Base & DoLa & OPERA & VCD & ICT & MaskCD & SID & \textbf{\modelname} \\
\midrule
\multirow{6}{*}{COCO}
 & \multirow{2}{*}{Random} & Acc & 85.23 & 86.53 & 87.31 & 88.63 & 87.53 & \underline{90.05} & 87.90 & \cellcolor{gray!12}\textbf{91.13} & 83.29 & 85.97 & \underline{89.20} & 87.73 & 89.18 & 88.55 & 88.05 & \cellcolor{gray!12}\textbf{89.27} \\
 &  & F1 & 83.09 & 84.44 & 86.92 & 87.81 & 85.84 & \underline{89.60} & 86.73 & \cellcolor{gray!12}\textbf{90.49} & 81.33 & 86.14 & \underline{88.81} & 87.16 & 88.48 & 88.30 & 87.34 & \cellcolor{gray!12}\textbf{89.18} \\
\cmidrule(l{2pt}){2-19}
 & \multirow{2}{*}{Popular} & Acc & 84.53 & 85.93 & 87.44 & 87.12 & 86.76 & \underline{88.65} & 86.90 & \cellcolor{gray!12}\textbf{89.70} & 81.88 & 82.93 & \underline{86.64} & 85.38 & 86.07 & 86.25 & 85.83 & \cellcolor{gray!12}\textbf{86.83} \\
 &  & F1 & 82.58 & 84.41 & 86.68 & 86.40 & 84.94 & \underline{88.45} & 86.00 & \cellcolor{gray!12}\textbf{89.12} & 80.06 & 83.80 & 84.89 & 85.06 & \underline{86.40} & 86.00 & 85.44 & \cellcolor{gray!12}\textbf{86.51} \\
\cmidrule(l{2pt}){2-19}
 & \multirow{2}{*}{Adversarial} & Acc & 83.37 & 83.47 & 84.78 & 84.26 & \underline{86.16} & 86.05 & 83.83 & \cellcolor{gray!12}\textbf{87.83} & 78.96 & 77.17 & 81.24 & 80.88 & \underline{83.30} & 81.80 & 81.41 & \cellcolor{gray!12}\textbf{84.07} \\
 &  & F1 & 81.57 & 81.51 & 83.45 & 83.90 & 84.32 & \underline{86.15} & 83.03 & \cellcolor{gray!12}\textbf{87.39} & 77.57 & 79.41 & 81.38 & 81.33 & \underline{82.54} & 82.10 & 81.54 & \cellcolor{gray!12}\textbf{83.41} \\
\midrule
\multirow{6}{*}{A-OKVQA}
 & \multirow{2}{*}{Random} & Acc & 86.40 & 87.40 & 88.19 & 89.22 & 88.96 & \underline{90.55} & 88.49 & \cellcolor{gray!12}\textbf{92.10} & 83.45 & 83.23 & 88.03 & 86.15 & \underline{89.00} & 87.45 & 86.47 & \cellcolor{gray!12}\textbf{89.13} \\
 &  & F1 & 85.07 & 86.46 & 88.43 & 89.01 & 87.90 & \underline{90.60} & 87.93 & \cellcolor{gray!12}\textbf{91.93} & 82.56 & 84.83 & 87.00 & 86.34 & \underline{88.71} & 87.55 & 86.52 & \cellcolor{gray!12}\textbf{89.52} \\
\cmidrule(l{2pt}){2-19}
 & \multirow{2}{*}{Popular} & Acc & 85.77 & 87.53 & 87.91 & 87.85 & 87.43 & \underline{89.05} & 87.62 & \cellcolor{gray!12}\textbf{89.67} & 79.90 & 76.47 & 83.22 & 81.85 & \underline{83.40} & 83.05 & 82.48 & \cellcolor{gray!12}\textbf{83.77} \\
 &  & F1 & 84.49 & 86.06 & 87.13 & 87.81 & 86.39 & \underline{89.15} & 87.41 & \cellcolor{gray!12}\textbf{89.70} & 79.59 & 79.86 & 84.15 & 82.82 & \underline{84.45} & 83.75 & 83.20 & \cellcolor{gray!12}\textbf{84.73} \\
\cmidrule(l{2pt}){2-19}
 & \multirow{2}{*}{Adversarial} & Acc & 80.37 & 81.83 & 80.82 & 81.27 & \underline{83.60} & 82.75 & 80.84 & \cellcolor{gray!12}\textbf{83.63} & 74.04 & 68.03 & 73.82 & 74.97 & 75.56 & \underline{75.90} & 75.50 & \cellcolor{gray!12}\textbf{77.30} \\
 &  & F1 & 79.68 & 81.33 & 81.54 & 82.38 & 83.02 & \underline{83.95} & 81.51 & \cellcolor{gray!12}\textbf{84.61} & 75.15 & 74.49 & 77.91 & 77.73 & \underline{78.68} & 78.40 & 77.94 & \cellcolor{gray!12}\textbf{79.43} \\
\midrule
\multirow{6}{*}{GQA}
 & \multirow{2}{*}{Random} & Acc & 85.10 & 87.13 & 86.02 & 85.59 & 88.96 & \underline{89.25} & 84.86 & \cellcolor{gray!12}\textbf{92.23} & 83.73 & 83.70 & 88.13 & 86.65 & \underline{89.23} & 87.85 & 86.97 & \cellcolor{gray!12}\textbf{89.47} \\
 &  & F1 & 83.87 & 86.05 & 85.29 & 85.33 & 87.89 & \underline{89.10} & 84.25 & \cellcolor{gray!12}\textbf{92.09} & 82.95 & 85.29 & 88.91 & 86.99 & \underline{89.34} & 88.25 & 87.17 & \cellcolor{gray!12}\textbf{89.53} \\
\cmidrule(l{2pt}){2-19}
 & \multirow{2}{*}{Popular} & Acc & 80.87 & 82.53 & 81.97 & 81.83 & \underline{86.43} & 86.35 & 81.61 & \cellcolor{gray!12}\textbf{88.37} & 78.17 & 74.03 & 79.27 & 80.73 & 80.86 & \underline{82.05} & 81.36 & \cellcolor{gray!12}\textbf{82.93} \\
 &  & F1 & 80.33 & 82.35 & 82.12 & 82.23 & 85.47 & \underline{86.70} & 81.83 & \cellcolor{gray!12}\textbf{88.60} & 78.37 & 78.75 & 82.11 & 82.24 & 82.64 & \underline{83.10} & 82.62 & \cellcolor{gray!12}\textbf{83.75} \\
\cmidrule(l{2pt}){2-19}
 & \multirow{2}{*}{Adversarial} & Acc & 78.77 & 82.00 & 80.24 & 80.01 & \underline{84.10} & 83.25 & 79.58 & \cellcolor{gray!12}\textbf{84.27} & 75.08 & 68.73 & 75.00 & 76.09 & \underline{77.40} & 77.10 & 76.62 & \cellcolor{gray!12}\textbf{79.60} \\
 &  & F1 & 78.56 & 81.51 & 80.64 & 80.75 & 83.53 & \underline{84.30} & 79.88 & \cellcolor{gray!12}\textbf{85.18} & 76.06 & 74.78 & 78.71 & 78.78 & \underline{80.11} & 79.30 & 78.99 & \cellcolor{gray!12}\textbf{81.17} \\
\bottomrule
\end{tabular}
}%
\caption{POPE results on COCO, A-OKVQA, and GQA. Bold denotes the best result, underline denotes the second-best result, and \modelname column is highlighted.}
\vspace{-8mm}
\label{tab:pope_full_llava_qwen25_delta}
\end{table}

Table~\ref{tab:pope_full_llava_qwen25_delta} reports the POPE results. 
\modelname achieves the best result on all 36 metrics across two backbones, three datasets, and three negative-sampling protocols. 
Averaged over all POPE settings, \modelname improves Qwen2.5-VL by $+5.4$ accuracy and $+6.7$ F1, and LLaVA-v1.5 by $+4.9$ accuracy and $+6.0$ F1 over their base decoders. 
It also consistently outperforms the strongest non-\modelname baseline in every setting, with average margins of $+1.23$ points on Qwen2.5-VL and $+0.63$ points on LLaVA-v1.5. 
The gains extend beyond accuracy: F1 improves across all splits, indicating that \modelname better balances rejecting absent objects and preserving correct positive predictions, rather than simply biasing the model toward negative answers. The advantage is stronger under greater language-prior pressure. 
On GQA, where visually absent but textually plausible objects make negative questions harder, Qwen2.5-VL improves from $80.9\%$ to $88.4\%$ accuracy on the Popular split and from $78.8\%$ to $84.3\%$ on the Adversarial split. 
LLaVA-v1.5 shows the same trend, increasing from $78.2\%$ to $82.9\%$ and from $75.1\%$ to $79.6\%$, respectively. 
This suggests that \modelname is especially effective when hallucinations arise from object co-occurrence priors rather than random decoding errors. 
While MaskCD and ICT are competitive in some settings, \modelname consistently surpasses them without external image perturbations, auxiliary intervention vectors, or duplicate full forward passes.

We further examine whether the POPE gains result from a simple Yes/No response-prior shift. 
Since POPE uses balanced positive and negative questions, uniformly changing the Yes-rate may improve one error type while hurting the other. 
\modelname does not exhibit this behavior: on LLaVA-v1.5 A-OKVQA-Adversarial and GQA-Adversarial, where the base decoder already has Yes-rates above $50\%$, \modelname further increases the Yes-rate by $+5.9$ and $+4.3$ points, yet accuracy still improves by $+3.3$ and $+4.5$ points and F1 by $+4.3$ and $+5.1$ points. 
Thus, the improvements cannot be explained by a uniform shift toward Yes or No responses, but are instead consistent with selective correction of visually grounded object decisions. 
Appendix~\ref{app:pope-decomp} provides the full per-split decomposition.

\subsection{CHAIR}
\label{sec:exp-chair}
Table~\ref{tab:chair-main} reports the CHAIR results on captioning. \modelname achieves the best results across all CHAIR settings, reducing both sentence-level and instance-level hallucination for the two backbones and generation lengths. On Qwen2.5-VL, CHAIRs drops from $20.8$ to $16.4$ at Len64 and from $35.5$ to $31.2$ at Len512; on LLaVA-v1.5, it drops from $31.5$ to $23.1$ and from $50.6$ to $43.6$, respectively. CHAIRi decreases jointly, e.g., from $17.8$ to $14.0$ on Qwen2.5-VL and from $20.3$ to $15.1$ on LLaVA-v1.5 at Len64, indicating that the gains do not come from simply omitting objects but from reducing spurious object mentions. The improvements remain consistent at Len512, suggesting that the internal contrast remains effective when hallucinations accumulate during longer autoregressive generation.

\subsection{AMBER}
\label{sec:exp-amber}

Table~\ref{tab:amber-main} reports the AMBER results. \modelname is best on all 8 metrics and, crucially, \emph{reduces hallucination while preserving or raising coverage}: on LLaVA-v1.5 Cover rises from $50.7$ to $53.7$ and Hal drops from $33.7$ to $21.6$, while on Qwen2.5-VL Cover is preserved ($47.2 \to 47.8$) as Hal falls from $23.6$ to $20.2$. This joint behavior is diagnostic, because other strong baselines obtain their Hal reductions at the expense of coverage: on LLaVA-v1.5, OPERA and VCD both drop Cover from $50.7$ to $49.6$ while lowering Hal, a pattern consistent with over-hedging or shortening the response rather than correcting the specific tokens that hallucinate.

The Cog metric, which captures cognitive hallucinations that are most tightly coupled to language priors, drops most sharply under \modelname ($3.8 \to 1.8$ on LLaVA-v1.5 and $1.9 \to 0.9$ on Qwen2.5-VL, each roughly halving). Because Cog isolates fabrications the model produces even when it is not certain the concept is in the image, its disproportionate reduction is precisely what we would expect from a mechanism that suppresses tokens driven by linguistic co-occurrence rather than by post-boundary visual evidence. \modelname thus removes the error type most aligned with its design target while preserving descriptive content, which we argue is the correct operating point for downstream use where over-hedged captions are as costly as hallucinated ones.

\begin{table}[!tbp]
\centering
\scriptsize
\setlength{\tabcolsep}{3.1pt}
\setlength{\aboverulesep}{0.25ex}
\setlength{\belowrulesep}{0.25ex}
\renewcommand{\arraystretch}{0.78}
\resizebox{\textwidth}{!}{%
\begin{tabular}{@{}cc|cc|cc|cccc@{}}
\toprule
\multirow{2}{*}{\textbf{Model}} & \multirow{2}{*}{\textbf{Method}} & \multicolumn{4}{c|}{\textbf{CHAIR}} & \multicolumn{4}{c@{}}{\textbf{AMBER}} \\
\cmidrule{3-6}\cmidrule{7-10}
& & $C_S^{64}\downarrow$ & $C_I^{64}\downarrow$ & $C_S^{512}\downarrow$ & $C_I^{512}\downarrow$ & CHAIR$\downarrow$ & Cover$\uparrow$ & Hal$\downarrow$ & Cog$\downarrow$ \\
\midrule
\multirow{9}{*}{Qwen2.5-VL}
& Baseline & 20.8 & 17.8 & 35.5 & 24.5 & 5.3 & 47.2 & 23.6 & 1.9 \\
& SID      & 18.1 & 15.6 & 33.0 & 22.8 & 4.9 & 46.3 & 21.5 & 1.4 \\
& MaskCD   & \underline{16.9} & 14.6 & \underline{31.8} & \underline{22.0} & 4.8 & \underline{47.4} & 22.0 & 1.5 \\
& DoLa     & 20.1 & 17.3 & 35.9 & 24.6 & 5.5 & 46.8 & 25.1 & 2.2 \\
& OPERA    & 18.6 & 15.9 & 32.9 & 22.8 & 5.1 & 46.2 & 21.6 & 1.7 \\
& VCD      & 19.3 & 16.7 & 33.9 & 23.3 & \underline{4.7} & 45.9 & \underline{20.5} & 1.6 \\
& ICT      & 17.0 & \underline{14.5} & 31.9 & 22.1 & 4.9 & 45.1 & 21.8 & \underline{1.2} \\
& \cellcolor{gray!12}\modelname (ours) & \cellcolor{gray!12}\textbf{16.4} & \cellcolor{gray!12}\textbf{14.0} & \cellcolor{gray!12}\textbf{31.2} & \cellcolor{gray!12}\textbf{21.7} & \cellcolor{gray!12}\textbf{4.6} & \cellcolor{gray!12}\textbf{47.8} & \cellcolor{gray!12}\textbf{20.2} & \cellcolor{gray!12}\textbf{0.9} \\
\midrule
\multirow{9}{*}{LLaVA-v1.5}
& Baseline & 31.5 & 20.3 & 50.6 & 30.4 & 7.3 & 50.7 & 33.7 & 3.8 \\
& SID      & 26.0 & 17.2 & 46.2 & 27.6 & 6.1 & 50.2 & 27.5 & 2.8 \\
& MaskCD   & \underline{24.0} & \underline{15.7} & \underline{44.2} & \underline{26.4} & 6.6 & \underline{52.4} & 31.6 & 3.0 \\
& DoLa     & 29.1 & 21.5 & 52.4 & 30.2 & 7.6 & 51.6 & 36.0 & 4.0 \\
& OPERA    & 26.7 & 17.9 & 46.9 & 27.8 & 7.3 & 49.6 & 32.0 & 3.5 \\
& VCD      & 28.3 & 19.0 & 48.7 & 28.9 & 6.8 & 49.6 & 30.4 & 3.5 \\
& ICT      & 24.2 & 15.9 & 44.5 & 26.6 & \underline{5.2} & 51.3 & \underline{23.1} & \underline{2.1} \\
& \cellcolor{gray!12}\modelname (ours) & \cellcolor{gray!12}\textbf{23.1} & \cellcolor{gray!12}\textbf{15.1} & \cellcolor{gray!12}\textbf{43.6} & \cellcolor{gray!12}\textbf{26.0} & \cellcolor{gray!12}\textbf{4.8} & \cellcolor{gray!12}\textbf{53.7} & \cellcolor{gray!12}\textbf{21.6} & \cellcolor{gray!12}\textbf{1.8} \\
\bottomrule
\end{tabular}
}
\caption{Main results on CHAIR and AMBER. CHAIR reports $C_S$/$C_I$ at two caption lengths (64, 512). For AMBER, CHAIR, Hal, and Cog are lower-is-better, while Cover is higher-is-better. Bold denotes the best result and underline denotes the second-best result.}
\vspace{-6mm}
\label{tab:chair-main}
\label{tab:amber-main}
\end{table}
\section{Discussion}
\label{sec:discussion}

\subsection{Boundary Placement for Internal Reconstruction}
\label{sec:disc-split}
\begin{figure}[h]
\centering
\includegraphics[width=\textwidth]{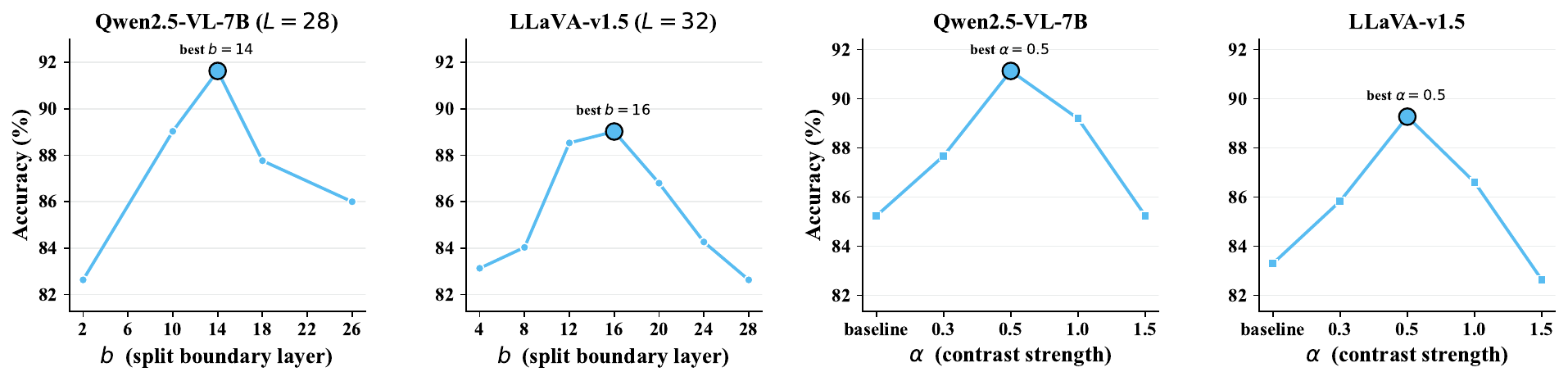}
\vspace{-4mm}
\caption{Effect of split boundary $b$ and contrast strength $\alpha$ on POPE accuracy for two models.}
\vspace{-4mm}
\label{fig:k-sweep-acc}
\end{figure}

\modelname's internal contrast is governed by the split boundary $b$ and contrast strength $\alpha$. Figure~\ref{fig:k-sweep-acc} sweeps both on POPE for Qwen2.5-VL and LLaVA-v1.5. Since $b=L-K$, moving $b$ changes the number of post-boundary layers in which the counterfactual branch is blocked from image-token access. Accuracy peaks at a middle split on both backbones and drops when the split is too early or too late. This supports the staged-fusion view in Section~\ref{sec:related-lvlm}: early splits disrupt multimodal grounding before a stable interpretation forms, whereas late splits leave little post-boundary computation for masking to change the prediction. A mid-layer boundary therefore provides the intended trade-off between preserving a shared multimodal prefix and retaining enough late layers for a meaningful internal contrast. The $\alpha$ sweep shows a similar balance: small $\alpha$ underuses the counterfactual reference, while large $\alpha$ over-amplifies the contrast and may penalize visually compatible tokens. Both backbones peak at $\alpha=0.5$, indicating that moderate prior subtraction best exposes language-template bias without eroding visually grounded predictions.

\begin{figure}[t]
\centering
\includegraphics[width=\textwidth]{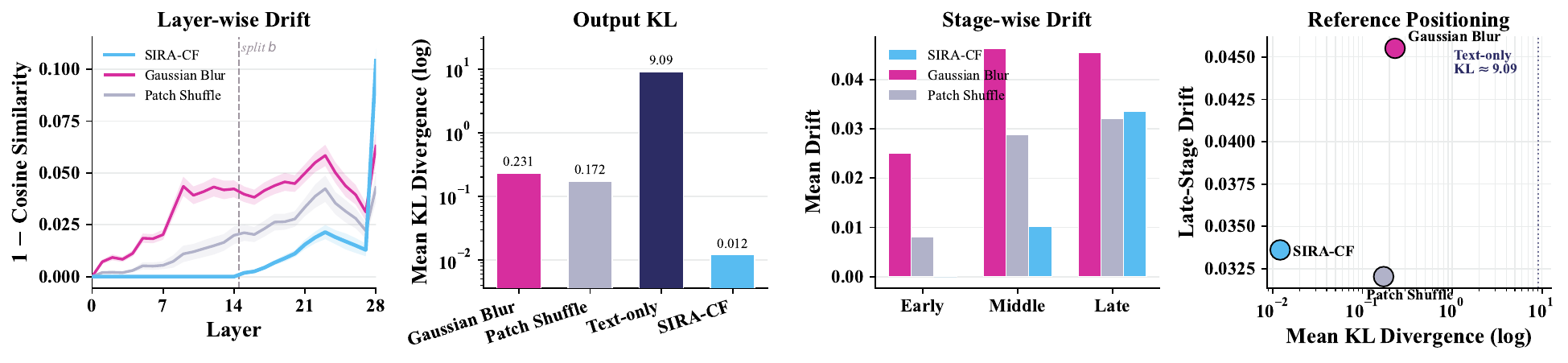}
\vspace{-3mm}
\caption{Contrastive reference analysis: (a) layer-wise drift; (b) next-token KL; (c) stage-wise drift; (d) KL versus late-stage drift.}
\vspace{-3mm}

\label{fig:contrastive-ref-analysis}
\end{figure}

\begin{figure}[h]
\centering
\includegraphics[width=\textwidth]{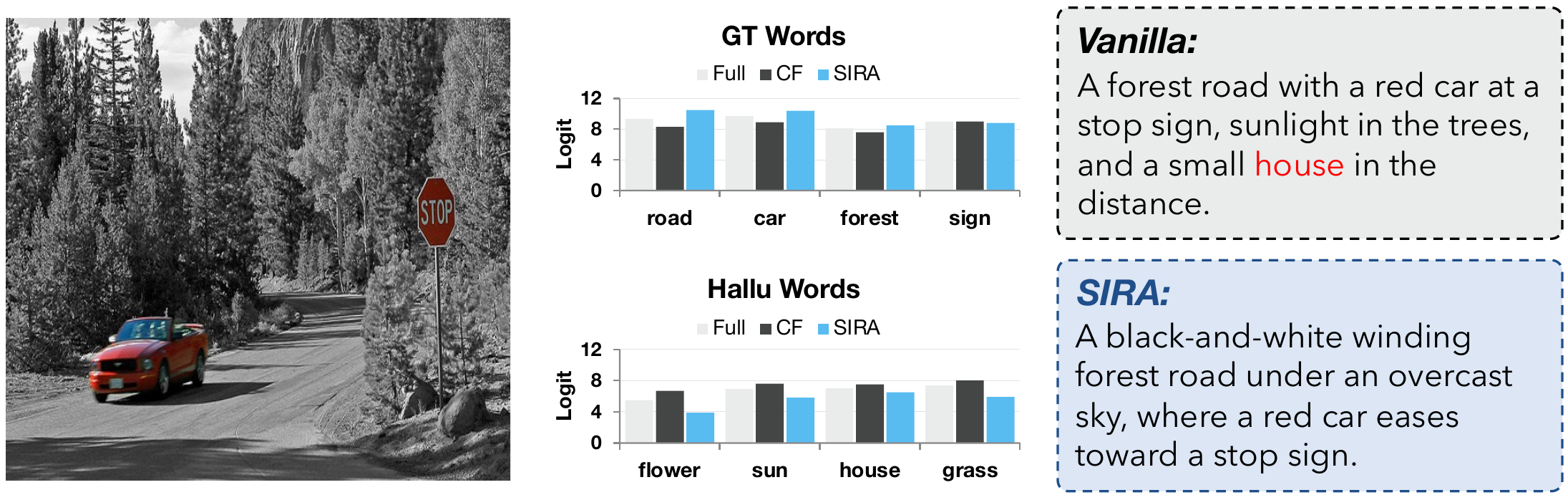}
\vspace{-5mm}

\caption{Case study and edge-case analysis on the AMBER benchmark using Qwen2.5-VL. Bars compare Full, CF, and \modelname logits for grounded tokens and hallucinated tokens.}
\label{fig:case-study}
\vspace{-5mm}
\end{figure}
\subsection{Alignment of the Internal Counterfactual Reference}
\label{sec:disc-ref-quality}

A useful counterfactual reference should preserve the original semantic trajectory before intervention and diverge primarily where visual access is intentionally removed. Figure~\ref{fig:contrastive-ref-analysis} shows that \modelname-CF satisfies this criterion better than input-space perturbations. Its layer-wise drift is nearly identical to the full branch before the boundary and grows mainly after the mask is applied, whereas Gaussian blur and patch shuffle drift throughout the network, including early layers that should preserve shared grounding. The output KL is also substantially lower for \modelname-CF ($0.012$) than for shuffle ($0.172$) or blur ($0.231$), while stage-wise statistics still show meaningful late-stage drift. This low-KL/high-late-drift pattern is desirable: a useful reference should not become a globally different generator, since a large output KL would make the logit gap reflect distributional mismatch or perturbation artifacts rather than visual attribution. At the same time, it should still deviate in the post-boundary layers where image-token access is removed, so that the resulting discrepancy can expose tokens that rely on continued visual evidence. Thus, \modelname-CF remains comparable to the full branch while separating at the intended intervention site, making it a better positioned internal reference for isolating visual contribution from language-prior bias.

\subsection{Contrast Strength and Token-Level Effect}
\label{sec:discussion-decoding}
\label{sec:disc-alpha}

Figure~\ref{fig:k-sweep-acc} shows the contrast is stable across a moderate range of $\alpha$, so we next inspect how it changes individual token scores. Figure~\ref{fig:case-study} shows the intended token-level effect of \modelname. For grounded objects, \modelname preserves strong logits under the full visual context. For hallucinated objects in the vanilla response, the counterfactual branch remains competitive, so \modelname lowers the final decoding logit and suppresses the hallucination. The same example also shows a subtle edge case. \modelname removes the unsupported object, but describes the scene as ``black-and-white''. This phrase is visually motivated because shadows fall across the road and make parts of the surface appear dark or desaturated. However, the road is not intrinsically black-and-white. The remaining issue is an attribute attribution error caused by shadow and lighting, rather than a clear language-prior object hallucination.

\FloatBarrier

\section{Conclusion and Limitations}
\label{sec:conclusion}

We presented \modelname, a training-free internal contrastive decoder for LVLM hallucination mitigation. By sharing early multimodal computation and branching only in late layers with image-token masking, SIRA constructs an aligned language-prior reference without external perturbations or a second full forward pass. Experiments on POPE, CHAIR, and AMBER show consistent hallucination reduction with preserved coverage. Its main limitations are the remaining late-layer decoding overhead and the requirement for white-box access to hidden states, masks, and caches.

\textbf{Limitations.} Because the post-boundary counterfactual branch is re-evaluated at every decoding step, \modelname also carries a per-step cost above standard decoding that, while well below VCD and OPERA (Table~\ref{tab:latency}), remains non-zero. Finally, constructing the internal contrast requires white-box access to hidden states, attention masks, and cache state, which restricts \modelname to open-weight or otherwise controllable LVLMs rather than black-box APIs.

\bibliographystyle{plainnat}
\bibliography{references}

@inproceedings{chen2025ict,
  title={Ict: Image-object cross-level trusted intervention for mitigating object hallucination in large vision-language models},
  author={Chen, Junzhe and Zhang, Tianshu and Huang, Shiyu and Niu, Yuwei and Zhang, Linfeng and Wen, Lijie and Hu, Xuming},
  booktitle={Proceedings of the Computer Vision and Pattern Recognition Conference},
  pages={4209--4221},
  year={2025}
}

@inproceedings{chuang2024dola,
  title={DoLa: Decoding by Contrasting Layers Improves Factuality in Large Language Models},
  author={Yung-Sung Chuang and Yujia Xie and Hongyin Luo and Yoon Kim and James R. Glass and Pengcheng He},
  booktitle={The Twelfth International Conference on Learning Representations},
  year={2024}
}

@inproceedings{leng2024mitigating,
  title={Mitigating object hallucinations in large vision-language models through visual contrastive decoding},
  author={Leng, Sicong and Zhang, Hang and Chen, Guanzheng and Li, Xin and Lu, Shijian and Miao, Chunyan and Bing, Lidong},
  booktitle={Proceedings of the IEEE/CVF Conference on Computer Vision and Pattern Recognition},
  pages={13872--13882},
  year={2024}
}

@inproceedings{huang2024opera,
  title={Opera: Alleviating hallucination in multi-modal large language models via over-trust penalty and retrospection-allocation},
  author={Huang, Qidong and Dong, Xiaoyi and Zhang, Pan and Wang, Bin and He, Conghui and Wang, Jiaqi and Lin, Dahua and Zhang, Weiming and Yu, Nenghai},
  booktitle={Proceedings of the IEEE/CVF Conference on Computer Vision and Pattern Recognition},
  pages={13418--13427},
  year={2024}
}

@inproceedings{li2023evaluating,
  title={Evaluating Object Hallucination in Large Vision-Language Models},
  author={Li, Yifan and Du, Yifan and Zhou, Kun and Wang, Jinpeng and Zhao, Wayne Xin and Wen, Ji-Rong},
  booktitle={Proceedings of the 2023 Conference on Empirical Methods in Natural Language Processing},
  pages={292--305},
  year={2023}
}

@inproceedings{rohrbach2018object,
  title={Object Hallucination in Image Captioning},
  author={Rohrbach, Anna and Hendricks, Lisa Anne and Burns, Kaylee and Darrell, Trevor and Saenko, Kate},
  booktitle={Proceedings of the 2018 Conference on Empirical Methods in Natural Language Processing},
  pages={4035--4045},
  year={2018}
}

@article{wang2023amber,
  title={Amber: An LLM-Free Multi-Dimensional Benchmark for MLLMs Hallucination Evaluation},
  author={Wang, Junyang and Wang, Yuhang and Xu, Guohai and Zhang, Jing and Gu, Yukai and Jia, Haitao and Wang, Jiaqi and Xu, Haiyang and Yan, Ming and Zhang, Ji and others},
  journal={arXiv preprint arXiv:2311.07397},
  year={2023}
}

@inproceedings{lin2014microsoft,
  title={Microsoft COCO: Common Objects in Context},
  author={Lin, Tsung-Yi and Maire, Michael and Belongie, Serge and Hays, James and Perona, Pietro and Ramanan, Deva and Doll{\'a}r, Piotr and Zitnick, C Lawrence},
  booktitle={European Conference on Computer Vision},
  pages={740--755},
  year={2014},
  organization={Springer}
}

@inproceedings{schwenk2022okvqa,
  title={A-OKVQA: A Benchmark for Visual Question Answering Using World Knowledge},
  author={Schwenk, Dustin and Khandelwal, Apoorv and Clark, Christopher and Marino, Kenneth and Mottaghi, Roozbeh},
  booktitle={European Conference on Computer Vision},
  pages={146--162},
  year={2022},
  organization={Springer}
}

@inproceedings{hudson2019gqa,
  title={GQA: A New Dataset for Real-World Visual Reasoning and Compositional Question Answering},
  author={Hudson, Drew A and Manning, Christopher D},
  booktitle={Proceedings of the IEEE/CVF Conference on Computer Vision and Pattern Recognition},
  pages={6700--6709},
  year={2019}
}

@misc{bai2025qwen25vltechnicalreport,
  title={Qwen2.5-VL Technical Report},
  author={Shuai Bai and Keqin Chen and Xuejing Liu and Jialin Wang and Wenbin Ge and Sibo Song and Kai Dang and Peng Wang and Shijie Wang and Jun Tang and Humen Zhong and Yuanzhi Zhu and Mingkun Yang and Zhaohai Li and Jianqiang Wan and Pengfei Wang and Wei Ding and Zheren Fu and Yiheng Xu and Jiabo Ye and Xi Zhang and Tianbao Xie and Zesen Cheng and Hang Zhang and Zhibo Yang and Haiyang Xu and Junyang Lin},
  year={2025},
  eprint={2502.13923},
  archivePrefix={arXiv},
  primaryClass={cs.CV},

}

@misc{liu2024improvedbaselinesvisualinstruction,
  title={Improved Baselines with Visual Instruction Tuning},
  author={Haotian Liu and Chunyuan Li and Yuheng Li and Yong Jae Lee},
  year={2024},
  eprint={2310.03744},
  archivePrefix={arXiv},
  primaryClass={cs.CV},

}

@inproceedings{yin2025lifting,
  author={Yin, Hao and Si, Guangzong and Wang, Zilei},
  booktitle={2025 IEEE/CVF Conference on Computer Vision and Pattern Recognition (CVPR)},
  title={Lifting the Veil on Visual Information Flow in MLLMs: Unlocking Pathways to Faster Inference},
  year={2025},
  pages={9382--9391},
  publisher={IEEE},
  doi={10.1109/CVPR52734.2025.00876}
}

@inproceedings{kaduri2025whats,
  author={Kaduri, Omri and Bagon, Shai and Dekel, Tali},
  booktitle={2025 IEEE/CVF Conference on Computer Vision and Pattern Recognition (CVPR)},
  title={What's in the Image? A Deep-Dive into the Vision of Vision Language Models},
  year={2025},
  pages={14549--14558},
  publisher={IEEE},
  doi={10.1109/CVPR52734.2025.01356}
}

@inproceedings{Ye_2024,
  title={mPLUG-OwI2: Revolutionizing Multi-modal Large Language Model with Modality Collaboration},
  author={Ye, Qinghao and Xu, Haiyang and Ye, Jiabo and Yan, Ming and Hu, Anwen and Liu, Haowei and Qian, Qi and Zhang, Ji and Huang, Fei},
  booktitle={2024 IEEE/CVF Conference on Computer Vision and Pattern Recognition (CVPR)},
  year={2024},
  pages={13040--13051},
  publisher={IEEE},
}

@inproceedings{Tong_2024,
  title={Eyes Wide Shut? Exploring the Visual Shortcomings of Multimodal LLMs},
  author={Tong, Shengbang and Liu, Zhuang and Zhai, Yuexiang and Ma, Yi and LeCun, Yann and Xie, Saining},
  booktitle={2024 IEEE/CVF Conference on Computer Vision and Pattern Recognition (CVPR)},
  year={2024},
  pages={9568--9578},
  publisher={IEEE},
}

@misc{meng2023locatingeditingfactualassociations,
  title={Locating and Editing Factual Associations in GPT},
  author={Kevin Meng and David Bau and Alex Andonian and Yonatan Belinkov},
  year={2022},
  eprint={2202.05262},
  archivePrefix={arXiv},
  primaryClass={cs.CL},
}

@article{Huang_2025,
  title={A Survey on Hallucination in Large Language Models: Principles, Taxonomy, Challenges, and Open Questions},
  author={Huang, Lei and Yu, Weijiang and Ma, Weitao and Zhong, Weihong and Feng, Zhangyin and Wang, Haotian and Chen, Qianglong and Peng, Weihua and Feng, Xiaocheng and Qin, Bing and Liu, Ting},
  journal={ACM Transactions on Information Systems},
  year={2025},
  volume={43},
  number={2},
  pages={1--55},
  publisher={Association for Computing Machinery (ACM)},
}

@inproceedings{Sun_2024,
  title={Aligning Large Multimodal Models with Factually Augmented RLHF},
  author={Sun, Zhiqing and Shen, Sheng and Cao, Shengcao and Liu, Haotian and Li, Chunyuan and Shen, Yikang and Gan, Chuang and Gui, Liangyan and Wang, Yu-Xiong and Yang, Yiming and Keutzer, Kurt and Darrell, Trevor},
  booktitle={Findings of the Association for Computational Linguistics ACL 2024},
  year={2024},
  pages={13088--13110},
  publisher={Association for Computational Linguistics},
}

@inproceedings{Yu_2024_2,
  title={RLHF-V: Towards Trustworthy MLLMs via Behavior Alignment from Fine-Grained Correctional Human Feedback},
  author={Yu, Tianyu and Yao, Yuan and Zhang, Haoye and He, Taiwen and Han, Yifeng and Cui, Ganqu and Hu, Jinyi and Liu, Zhiyuan and Zheng, Hai-Tao and Sun, Maosong},
  booktitle={2024 IEEE/CVF Conference on Computer Vision and Pattern Recognition (CVPR)},
  year={2024},
  pages={13807--13816},
  publisher={IEEE},
}

@inproceedings{Yu_2024,
  title={HalluciDoctor: Mitigating Hallucinatory Toxicity in Visual Instruction Data},
  author={Yu, Qifan and Li, Juncheng and Wei, Longhui and Pang, Liang and Ye, Wentao and Qin, Bosheng and Tang, Siliang and Tian, Qi and Zhuang, Yueting},
  booktitle={2024 IEEE/CVF Conference on Computer Vision and Pattern Recognition (CVPR)},
  year={2024},
  pages={12944--12953},
  publisher={IEEE},
}

@inbook{Zhang_2024_2,
  title={Reflective Instruction Tuning: Mitigating Hallucinations in Large Vision-Language Models},
  author={Zhang, Jinrui and Wang, Teng and Zhang, Haigang and Lu, Ping and Zheng, Feng},
  booktitle={Computer Vision -- ECCV 2024},
  year={2024},
  pages={196--213},
  publisher={Springer Nature Switzerland},
}

@article{Yin_2024,
  title={Woodpecker: hallucination correction for multimodal large language models},
  author={Yin, Shukang and Fu, Chaoyou and Zhao, Sirui and Xu, Tong and Wang, Hao and Sui, Dianbo and Shen, Yunhang and Li, Ke and Sun, Xing and Chen, Enhong},
  journal={Science China Information Sciences},
  year={2024},
  volume={67},
  number={12},
  publisher={Springer Science and Business Media LLC},
}

@inproceedings{Lee_2025,
  title={VLind-Bench: Measuring Language Priors in Large Vision-Language Models},
  author={Lee, Kang-il and Kim, Minbeom and Yoon, Seunghyun and Kim, Minsung and Lee, Dongryeol and Koh, Hyukhun and Jung, Kyomin},
  booktitle={Findings of the Association for Computational Linguistics: NAACL 2025},
  year={2025},
  pages={4129--4144},
  publisher={Association for Computational Linguistics},
}

@inproceedings{Zhong_2024,
  title={Investigating and Mitigating the Multimodal Hallucination Snowballing in Large Vision-Language Models},
  author={Zhong, Weihong and Feng, Xiaocheng and Zhao, Liang and Li, Qiming and Huang, Lei and Gu, Yuxuan and Ma, Weitao and Xu, Yuan and Qin, Bing},
  booktitle={Proceedings of the 62nd Annual Meeting of the Association for Computational Linguistics (Volume 1: Long Papers)},
  year={2024},
  pages={11991--12011},
  publisher={Association for Computational Linguistics},
}

@article{Gunjal_2024,
  title={Detecting and Preventing Hallucinations in Large Vision Language Models},
  author={Gunjal, Anisha and Yin, Jihan and Bas, Erhan},
  journal={Proceedings of the AAAI Conference on Artificial Intelligence},
  year={2024},
  volume={38},
  number={16},
  pages={18135--18143},
  publisher={Association for the Advancement of Artificial Intelligence (AAAI)},
}

@inproceedings{Jiang_2024,
  title={Hallucination Augmented Contrastive Learning for Multimodal Large Language Model},
  author={Jiang, Chaoya and Xu, Haiyang and Dong, Mengfan and Chen, Jiaxing and Ye, Wei and Yan, Ming and Ye, Qinghao and Zhang, Ji and Huang, Fei and Zhang, Shikun},
  booktitle={2024 IEEE/CVF Conference on Computer Vision and Pattern Recognition (CVPR)},
  year={2024},
  pages={27026--27036},
  publisher={IEEE},
}

@inproceedings{Li_2023,
  title={Contrastive Decoding: Open-ended Text Generation as Optimization},
  author={Li, Xiang Lisa and Holtzman, Ari and Fried, Daniel and Liang, Percy and Eisner, Jason and Hashimoto, Tatsunori and Zettlemoyer, Luke and Lewis, Mike},
  booktitle={Proceedings of the 61st Annual Meeting of the Association for Computational Linguistics (Volume 1: Long Papers)},
  year={2023},
  pages={12286--12312},
  publisher={Association for Computational Linguistics},
}

@inproceedings{Kim_2024,
  title={CODE: Contrasting Self-generated Description to Combat Hallucination in Large Multi-modal Models},
  author={Kim, Hyun and Kim, Junho and Kim, Yeon and Ro, Yong},
  booktitle={Advances in Neural Information Processing Systems 37},
  year={2024},
  pages={133571--133599},
  publisher={Neural Information Processing Systems Foundation, Inc. (NeurIPS)},
}

@inproceedings{An_2025,
  title={Mitigating Object Hallucinations in Large Vision-Language Models with Assembly of Global and Local Attention},
  author={An, Wenbin and Tian, Feng and Leng, Sicong and Nie, Jiahao and Lin, Haonan and Wang, Qianying and Chen, Ping and Zhang, Xiaoqin and Lu, Shijian},
  booktitle={2025 IEEE/CVF Conference on Computer Vision and Pattern Recognition (CVPR)},
  year={2025},
  pages={29915--29926},
  publisher={IEEE},
}

@inproceedings{Woo_2025,
  title={Don't Miss the Forest for the Trees: Attentional Vision Calibration for Large Vision Language Models},
  author={Woo, Sangmin and Kim, Donguk and Jang, Jaehyuk and Choi, Yubin and Kim, Changick},
  booktitle={Findings of the Association for Computational Linguistics: ACL 2025},
  year={2025},
  pages={1927--1951},
  publisher={Association for Computational Linguistics},
}

@inbook{Liu_2024_3,
  title={Paying More Attention to Image: A Training-Free Method for Alleviating Hallucination in LVLMs},
  author={Liu, Shi and Zheng, Kecheng and Chen, Wei},
  booktitle={Computer Vision -- ECCV 2024},
  year={2024},
  pages={125--140},
  publisher={Springer Nature Switzerland},
}

@inproceedings{Favero_2024,
  title={Multi-Modal Hallucination Control by Visual Information Grounding},
  author={Favero, Alessandro and Zancato, Luca and Trager, Matthew and Choudhary, Siddharth and Perera, Pramuditha and Achille, Alessandro and Swaminathan, Ashwin and Soatto, Stefano},
  booktitle={2024 IEEE/CVF Conference on Computer Vision and Pattern Recognition (CVPR)},
  year={2024},
  pages={14303--14312},
  publisher={IEEE},
}

@inproceedings{Yue_2024,
  title={Less is More: Mitigating Multimodal Hallucination from an EOS Decision Perspective},
  author={Yue, Zihao and Zhang, Liang and Jin, Qin},
  booktitle={Proceedings of the 62nd Annual Meeting of the Association for Computational Linguistics (Volume 1: Long Papers)},
  year={2024},
  pages={11766--11781},
  publisher={Association for Computational Linguistics},
}

@misc{touvron2023llama,
  title={LLaMA: Open and Efficient Foundation Language Models},
  author={Hugo Touvron and Thibaut Lavril and Gautier Izacard and Xavier Martinet and Marie-Anne Lachaux and Timoth{\'e}e Lacroix and Baptiste Rozi{\`e}re and Naman Goyal and Eric Hambro and Faisal Azhar and Aurelien Rodriguez and Armand Joulin and Edouard Grave and Guillaume Lample},
  year={2023},
  eprint={2302.13971},
  archivePrefix={arXiv},
  primaryClass={cs.CL},

}

@article{touvron2023llama2,
  title={Llama 2: Open foundation and fine-tuned chat models},
  author={Touvron, Hugo and Martin, Louis and Stone, Kevin and Albert, Peter and Almahairi, Amjad and Babaei, Yasmine and Bashlykov, Nikolay and Batra, Soumya and Bhargava, Prajjwal and Bhosale, Shruti and others},
  journal={arXiv preprint arXiv:2307.09288},
  year={2023}
}

@article{achiam2023gpt4,
  title={GPT-4 technical report},
  author={Achiam, Josh and Adler, Steven and Agarwal, Sandhini and Ahmad, Lama and Akkaya, Ilge and Aleman, Florencia Leoni and Almeida, Diogo and Altenschmidt, Janko and Altman, Sam and Anadkat, Shyamal and others},
  journal={arXiv preprint arXiv:2303.08774},
  year={2023}
}

@misc{li2023blip2,
  title={BLIP-2: Bootstrapping Language-Image Pre-training with Frozen Image Encoders and Large Language Models},
  author={Junnan Li and Dongxu Li and Silvio Savarese and Steven Hoi},
  year={2023},
  eprint={2301.12597},
  archivePrefix={arXiv},
  primaryClass={cs.CV},

}

@misc{geminiteam2025gemini,
  title={Gemini: A Family of Highly Capable Multimodal Models},
  author={Gemini Team and Rohan Anil and Sebastian Borgeaud and Jean-Baptiste Alayrac and Jiahui Yu and Radu Soricut and Johan Schalkwyk and Andrew M. Dai and Anja Hauth and Katie Millican and David Silver and Melvin Johnson and Ioannis Antonoglou and Julian Schrittwieser and Amelia Glaese and Jilin Chen and Emily Pitler and Timothy Lillicrap and Angeliki Lazaridou and Orhan Firat and others},
  year={2025},
  eprint={2312.11805},
  archivePrefix={arXiv},
  primaryClass={cs.CL},

}

@inproceedings{yin2026the,
  title={The Mirage of Performance Gains: Why Contrastive Decoding Fails to Mitigate Object Hallucinations in {MLLM}s?},
  author={Hao Yin and Guangzong Si and Zilei Wang},
  booktitle={The Thirty-ninth Annual Conference on Neural Information Processing Systems},
  year={2026},
}

@inproceedings{wang-etal-2024-mitigating,
    title = "Mitigating Hallucinations in Large Vision-Language Models with Instruction Contrastive Decoding",
    author = "Wang, Xintong  and Pan, Jingheng  and Ding, Liang  and Biemann, Chris",
    booktitle = "Findings of the Association for Computational Linguistics ACL 2024",
    year = "2024",
    pages = "15840--15853",
}

@inproceedings{deng-yang-2025-maskcd,
    title = "{M}ask{CD}: Mitigating {LVLM} Hallucinations by Image Head Masked Contrastive Decoding",
    author = "Deng, Jingyuan  and
      Yang, Yujiu",
    editor = "Christodoulopoulos, Christos  and
      Chakraborty, Tanmoy  and
      Rose, Carolyn  and
      Peng, Violet",
    booktitle = "Findings of the Association for Computational Linguistics: EMNLP 2025",
    month = nov,
    year = "2025",
    address = "Suzhou, China",
    publisher = "Association for Computational Linguistics",
    doi = "10.18653/v1/2025.findings-emnlp.1025",
    pages = "18854--18866",
    ISBN = "979-8-89176-335-7"
}

@inproceedings{liu-etal-2025-multi,
    title = "Multi-Frequency Contrastive Decoding: Alleviating Hallucinations for Large Vision-Language Models",
    author = "Liu, Bingqian  and
      Zhang, Fu  and
      Chen, Guoqing  and
      Cheng, Jingwei",
    editor = "Christodoulopoulos, Christos  and
      Chakraborty, Tanmoy  and
      Rose, Carolyn  and
      Peng, Violet",
    booktitle = "Proceedings of the 2025 Conference on Empirical Methods in Natural Language Processing",
    month = nov,
    year = "2025",
    address = "Suzhou, China",
    publisher = "Association for Computational Linguistics",
    doi = "10.18653/v1/2025.emnlp-main.1452",
    pages = "28568--28584",
    ISBN = "979-8-89176-332-6"
}

@misc{tong2025mitigatinghallucinationmultimodalllms,
      title={Mitigating Hallucination in Multimodal LLMs with Layer Contrastive Decoding}, 
      author={Bingkui Tong and Jiaer Xia and Kaiyang Zhou},
      year={2025},
      eprint={2509.25177},
      archivePrefix={arXiv},
      primaryClass={cs.CV},
}

@inproceedings{chen-etal-2026-mask,
    title = "Mask What Matters: Mitigating Object Hallucinations in Multimodal Large Language Models with Object-Aligned Visual Contrastive Decoding",
    author = "Chen, Boqi  and
      Liu, Xudong  and
      Qiu, Jianing",
    editor = "Baez Santamaria, Selene  and
      Somayajula, Sai Ashish  and
      Yamaguchi, Atsuki",
    booktitle = "Proceedings of the 19th Conference of the {E}uropean Chapter of the {A}ssociation for {C}omputational {L}inguistics (Volume 4: Student Research Workshop)",
    month = mar,
    year = "2026",
    address = "Rabat, Morocco",
    publisher = "Association for Computational Linguistics",
    pages = "9--16",
    ISBN = "979-8-89176-383-8"
}

@inproceedings{wan2025only,
  title={Only: One-layer intervention sufficiently mitigates hallucinations in large vision-language models},
  author={Wan, Zifu and Zhang, Ce and Yong, Silong and Ma, Martin Q and Stepputtis, Simon and Morency, Louis-Philippe and Ramanan, Deva and Sycara, Katia and Xie, Yaqi},
  booktitle={Proceedings of the IEEE/CVF International Conference on Computer Vision},
  pages={3225--3234},
  year={2025}
}

@inproceedings{huo2025selfintrospective,
  title={Self-Introspective Decoding: Alleviating Hallucinations for Large Vision-Language Models},
  author={Fushuo Huo and Wenchao Xu and Zhong Zhang and Haozhao Wang and Zhicheng Chen and Peilin Zhao},
  booktitle={The Thirteenth International Conference on Learning Representations},
  year={2025},
}

\appendix
\section{Implementation Details}
\label{app:imp}
All methods are evaluated on identical prompts and image inputs, with each baseline run under its original decoding configuration. For \modelname we use $\alpha=0.5$ with $K=14$ on Qwen2.5-VL and $K=16$ on LLaVA-v1.5, applying the counterfactual mask from Section~\ref{sec:method-cf} without modification; hyperparameters are fixed across datasets. All experiments are conducted on a system equipped with 3 NVIDIA A100 40GB GPUs.
\section{\modelname Decoding Algorithm}
\label{app:algo}

\begin{algorithm}[ht]
\caption{\modelname: Single-Pass Internal Contrastive Decoding}
\label{alg:sira}
\begin{algorithmic}[1]
\Require LVLM with shared prefix layers $F_{0:b-1}$, post-boundary layers $F_{b:L-1}$, normalization $\mathrm{Norm}(\cdot)$, output projection $W_{\mathrm{out}}$; prompt $\mathbf{x}_{1:S}$; image-token positions $\mathcal{P}_{\mathrm{img}} \subseteq \{1, \ldots, S\}$; boundary $b = L - K$; strength $\alpha \geq 0$; max length $T$
\Ensure generated sequence $\mathbf{y}$
\State $\mathbf{h}^{b}_{1},\; C^{\text{shared}}_{0:b} \gets F_{0:b-1}(\mathbf{x}_{1:S})$ \Comment{shared prefix, single pass}
\State $C^{\text{full}}_{0:b},\; C^{\text{cf}}_{0:b} \gets C^{\text{shared}}_{0:b}$
\State $M^{\mathrm{causal}}_{1} \gets$ standard causal mask of shape $S \times S$
\State $M^{\mathrm{cf}}_{1} \gets M^{\mathrm{causal}}_{1}$ with rows $q \in \mathcal{P}_{\mathrm{img}}$ and columns $k \in \mathcal{P}_{\mathrm{img}}$ set to $-\infty$
\State $\mathbf{h}^{\mathrm{full},L}_{1},\; C^{\text{full}}_{b:L} \gets F_{b:L-1}\!\left(\mathbf{h}^{b}_{1};\, M^{\mathrm{causal}}_{1}\right)$
\State $\mathbf{h}^{\mathrm{cf},L}_{1},\; C^{\text{cf}}_{b:L} \gets F_{b:L-1}\!\left(\mathbf{h}^{b}_{1};\, M^{\mathrm{cf}}_{1}\right)$
\State $z^{\mathrm{full}}_{1} \gets W_{\mathrm{out}}\,\mathrm{Norm}\!\left(\mathbf{h}^{\mathrm{full},L}_{1}\right)_{\mathrm{last}}$;\quad $z^{\mathrm{cf}}_{1} \gets W_{\mathrm{out}}\,\mathrm{Norm}\!\left(\mathbf{h}^{\mathrm{cf},L}_{1}\right)_{\mathrm{last}}$ \Comment{float32}
\State $z^{\mathrm{cd}}_{1} \gets (1{+}\alpha)\, z^{\mathrm{full}}_{1} - \alpha\, z^{\mathrm{cf}}_{1}$;\quad $y_{1} \gets \arg\max_{v \in \mathcal{V}}\, z^{\mathrm{cd}}_{1}(v)$;\quad $\mathbf{y} \gets (y_{1})$
\For{$t = 2, \ldots, T$}
  \State extend $M^{\mathrm{causal}}_{t}, M^{\mathrm{cf}}_{t}$ by one new query row at position $S{+}t{-}1$; in $M^{\mathrm{cf}}_{t}$ set columns $k \in \mathcal{P}_{\mathrm{img}}$ to $-\infty$
  \State $\mathbf{h}^{b}_{t},\; C^{\text{full}}_{0:b} \gets F_{0:b-1}\!\left(y_{t-1};\, C^{\text{full}}_{0:b}\right)$ \Comment{shared, single-token forward}
  \State $C^{\text{cf}}_{0:b} \gets C^{\text{full}}_{0:b}$ \Comment{shared layers stay identical}
  \State $\mathbf{h}^{\mathrm{full},L}_{t},\; C^{\text{full}}_{b:L} \gets F_{b:L-1}\!\left(\mathbf{h}^{b}_{t};\, C^{\text{full}}_{b:L}, M^{\mathrm{causal}}_{t}\right)$
  \State $\mathbf{h}^{\mathrm{cf},L}_{t},\; C^{\text{cf}}_{b:L} \gets F_{b:L-1}\!\left(\mathbf{h}^{b}_{t};\, C^{\text{cf}}_{b:L}, M^{\mathrm{cf}}_{t}\right)$
  \State $z^{\mathrm{full}}_{t} \gets W_{\mathrm{out}}\,\mathrm{Norm}\!\left(\mathbf{h}^{\mathrm{full},L}_{t}\right)_{\mathrm{last}}$;\quad $z^{\mathrm{cf}}_{t} \gets W_{\mathrm{out}}\,\mathrm{Norm}\!\left(\mathbf{h}^{\mathrm{cf},L}_{t}\right)_{\mathrm{last}}$ \Comment{float32}
  \State $z^{\mathrm{cd}}_{t} \gets (1{+}\alpha)\, z^{\mathrm{full}}_{t} - \alpha\, z^{\mathrm{cf}}_{t}$;\quad $y_{t} \gets \arg\max_{v \in \mathcal{V}}\, z^{\mathrm{cd}}_{t}(v)$;\quad append $y_{t}$ to $\mathbf{y}$
  \If{$y_{t}$ is the end-of-sequence token} \textbf{break} \EndIf
\EndFor
\State \Return $\mathbf{y}$
\end{algorithmic}
\end{algorithm}

The notation reuses Section~\ref{sec:method}: $F_{0:b-1}, F_{b:L-1}$ are the shared prefix and post-boundary layer stacks; $M^{\mathrm{causal}}_{t}, M^{\mathrm{cf}}_{t}$ are the causal and counterfactual attention masks, where $M^{\mathrm{cf}}_{t}$ blocks every entry whose query row or key column lies in $\mathcal{P}_{\mathrm{img}}$; $\mathrm{Norm}(\cdot)$ and $W_{\mathrm{out}}$ are the backbone's final normalization and output projection. We additionally write $C^{*}_{a:b}$ for branch ${*}$'s key/value cache over layers $a, \ldots, b{-}1$, and $F_{a:b}(\cdot;\, C, M)$ for the cached forward that consumes $C$, appends new key/value rows, and returns the resulting layer-$b$ hidden states together with the updated cache. Lines~1--2 share the prefix cache between branches without recomputation; lines~3--4 build the two prompt-time masks of shape $S \times S$; lines~5--8 perform the late-layer prefill that yields the first generated token $y_{1}$. The decoding loop runs the shared layers $F_{0:b-1}$ \emph{once} per step on the previous token (line~11) and forks only at layer $b$, so \modelname's per-step overhead is $K$ layers of attention and MLP on the last token, giving a latency ratio of approximately $1 + K/L$ rather than the $2\times$ of VCD (Table~\ref{tab:latency}).

Two implementation notes are worth recording. First, the contrasts in lines~7 and~15 are computed in float32 to eliminate half-precision rank reversals on the tail of the vocabulary when the backbone runs in bfloat16. Second, the CF masking \emph{rule}---block keys (and, during prefill, also queries) at $\mathcal{P}_{\mathrm{img}}$ in the CF branch only---is fixed at request time, so only one mask row is appended per decoding step (line~10) and no per-request mask reconstruction is needed. No adaptive plausibility constraint or vocabulary truncation is applied: as argued in Section~\ref{sec:related-hallucination}, the CF distribution already lives on the base model's token manifold.

\section{Baseline Details and Comparison}
\label{app:baseline-details}

We compare against representative inference-time hallucination mitigation methods. DoLa~\citep{chuang2024dola} contrasts logits from different transformer layers to favor predictions that are more strongly supported by mature representations. VCD~\citep{leng2024mitigating} constructs a visual contrast by comparing the original image with a distorted visual input, such as a perturbed or degraded image. OPERA~\citep{huang2024opera} penalizes over-trusted attention patterns and uses retrospection allocation during generation. ICT~\citep{chen2025ict} introduces image-object cross-level trusted intervention to steer decoding with object-level signals. MaskCD~\citep{deng-yang-2025-maskcd} masks image heads to form a contrastive decoding signal. SID~\citep{huo2025selfintrospective} uses self-introspective decoding to construct an additional reference for hallucination reduction.

Table~\ref{tab:baseline-properties} summarizes method-level properties and trade-offs rather than numerical performance. We organize the comparison around four practical questions: whether the method is tailored to LVLMs, preserves the original visual input, avoids external tools/models, and keeps decoding overhead low. Several baselines meet part of these requirements, while \modelname satisfies all four.

\begin{table}[!ht]
\centering
\footnotesize
\setlength{\tabcolsep}{5.2pt}
\renewcommand{\arraystretch}{1.18}
\begin{tabular*}{\textwidth}{@{\extracolsep{\fill}}lcccc@{}}
\toprule
Method & LVLM-specific & Preserves input & No external tools/models & Low decoding overhead \\
\midrule
DoLa~\citep{chuang2024dola} & \xmark & \cmark & \cmark & \cmark \\
VCD~\citep{leng2024mitigating} & \cmark & \xmark & \cmark & \xmark \\
OPERA~\citep{huang2024opera} & \cmark & \cmark & \cmark & \xmark \\
ICT~\citep{chen2025ict} & \cmark & \cmark & \xmark & \cmark \\
MaskCD~\citep{deng-yang-2025-maskcd} & \cmark & \cmark & \cmark & \xmark \\
SID~\citep{huo2025selfintrospective} & \cmark & \cmark & \cmark & \xmark \\
\midrule
\addlinespace[1pt]
\textbf{\modelname (ours)} & \textbf{\cmark} & \textbf{\cmark} & \textbf{\cmark} & \textbf{\cmark} \\
\bottomrule
\end{tabular*}
\caption{Property comparison of representative inference-time hallucination mitigation methods. Low decoding overhead means avoiding extra full reference passes, self-introspective reference construction, or beam-search retrospection.}
\label{tab:baseline-properties}
\end{table}

\section{Efficiency Analysis}
\label{app:efficiency}

\modelname's efficiency follows directly from shared-prefix branching. A standard autoregressive step evaluates all $L$ layers once. \modelname evaluates the same full branch while adding only a counterfactual continuation through the final $K$ post-boundary layers, so the per-step compute scales as $L+K$ layers and the idealized ratio is $(L+K)/L = 1 + K/L$, rather than two full $L$-layer passes. Our measurements below focus on wall-clock latency.

Table~\ref{tab:latency} quantifies the practical overhead. \modelname adds $1.4$--$1.5\times$ overhead over standard decoding, compared with $\sim\!2.1\times$ for VCD, which requires a full duplicate pass on a perturbed image, and $5$--$6\times$ for OPERA's beam-search manipulation. The ratio remains stable as generation length grows from 32 to 128 tokens, because the per-step cost is dominated by late-layer branching rather than by a second full multimodal trajectory.

\begin{table}[!ht]
\centering
\scriptsize
\setlength{\tabcolsep}{3pt}
\renewcommand{\arraystretch}{1.05}
\newcommand{\latred}[1]{\textcolor{red}{\raisebox{0.05ex}{\scalebox{0.78}[0.88]{$\uparrow$}}\,\ensuremath{\times #1}}}
\newcommand{\latgreen}[1]{\textcolor{green!50!black}{\raisebox{0.05ex}{\scalebox{0.78}[0.88]{$\uparrow$}}\,\ensuremath{\times #1}}}
\begin{tabular*}{\textwidth}{@{\extracolsep{\fill}}lrcrcrc@{}}
\toprule
Method & \multicolumn{2}{c}{32-Token Len} & \multicolumn{2}{c}{64-Token Len} & \multicolumn{2}{c}{128-Token Len} \\
\cmidrule(lr){2-3}\cmidrule(lr){4-5}\cmidrule(lr){6-7}
 & ms & Rel. & ms & Rel. & ms & Rel. \\
\midrule
Qwen2.5-VL & 587 & $\times$1.0 & 1000 & $\times$1.0 & 1868 & $\times$1.0 \\
VCD & 1205 & \latred{2.1} & 2063 & \latred{2.1} & 3847 & \latred{2.1} \\
OPERA & 3100 & \latred{5.3} & 7786 & \latred{5.8} & 10966 & \latred{5.9} \\
\midrule
\modelname (ours) & \textbf{838} & \textbf{\latgreen{1.4}} & \textbf{1496} & \textbf{\latgreen{1.5}} & \textbf{2781} & \textbf{\latgreen{1.5}} \\
\bottomrule
\end{tabular*}
\caption{Comparison of inference efficiency across methods on Qwen2.5-VL with varying generation lengths. Inference times are recorded in milliseconds.}
\label{tab:latency}
\end{table}

\section{POPE Yes-Rate Decomposition}
\label{app:pope-decomp}

Table~\ref{tab:pope-decomp-qwen25vl} and Table~\ref{tab:pope-decomp-llavav15} report the full POPE decomposition (Accuracy, F1, Precision, Recall, FPR, Yes-rate) for every (backbone, dataset, split, method) cell. Across both backbones and all nine splits, \modelname's Yes-rate rises relative to the backbone (by $+2.2$ to $+9.8$ points), so Yes-rate shift alone is not the distinguishing quantity. The sharper diagnostic is how Yes-rate moves \emph{with respect to accuracy}: on POPE splits whose baseline Yes-rate already exceeds $50\%$, a pure Yes/No rebalancing would push the decoder past the balanced optimum and \emph{degrade} accuracy. \modelname exhibits the opposite behavior on every such split: on LLaVA-v1.5 A-OKVQA-Adversarial (baseline Yes $54.5\%$, \modelname Yes $60.4\%$) accuracy rises from $74.0$ to $77.3$, and on LLaVA-v1.5 GQA-Adversarial (baseline Yes $54.1\%$, \modelname Yes $58.3\%$) accuracy rises from $75.1$ to $79.6$. The Yes-rate increase therefore does not capture a unidirectional rebalancing of Yes/No mass but accompanies genuine visually grounded correction.

\begin{table}[!ht]
\centering
\scriptsize
\setlength{\tabcolsep}{4.0pt}
\renewcommand{\arraystretch}{0.92}
\begin{tabular*}{\textwidth}{@{\extracolsep{\fill}}lllrrrrrr@{}}
\toprule
Dataset & Setting & Method & Acc & F1 & Prec.\,\% & Recall\,\% & FPR\,\% & Yes\,\% \\
\midrule
\multirow{24}{*}{COCO} & \multirow{8}{*}{Random} & Baseline & 85.23 & 83.09 & 97.17 & 72.57 & 2.11 & 37.34 \\
 &  & DoLa & 86.53 & 84.44 & 99.95 & 73.10 & 0.04 & 36.57 \\
 &  & OPERA & 87.31 & 86.92 & 89.68 & 84.33 & 9.71 & 47.02 \\
 &  & VCD & 88.63 & 87.81 & 94.64 & 81.90 & 4.64 & 43.27 \\
 &  & ICT & 87.53 & 85.84 & 99.30 & 75.59 & 0.53 & 38.06 \\
 &  & MaskCD & 90.05 & 89.60 & 93.84 & 85.72 & 5.62 & 45.67 \\
 &  & SID & 87.90 & 86.73 & 96.01 & 79.08 & 3.28 & 41.18 \\
 &  & \modelname & 91.13 & 90.49 & 97.53 & 84.40 & 2.14 & 43.27 \\
\cmidrule(l){2-9}
 & \multirow{8}{*}{Popular} & Baseline & 84.53 & 82.58 & 94.49 & 73.34 & 4.28 & 38.81 \\
 &  & DoLa & 85.93 & 84.41 & 94.63 & 76.18 & 4.32 & 40.25 \\
 &  & OPERA & 87.44 & 86.68 & 92.26 & 81.73 & 6.85 & 44.29 \\
 &  & VCD & 87.12 & 86.40 & 91.52 & 81.83 & 7.59 & 44.71 \\
 &  & ICT & 86.76 & 84.94 & 98.48 & 74.68 & 1.16 & 37.92 \\
 &  & MaskCD & 88.65 & 88.45 & 90.04 & 86.92 & 9.62 & 48.27 \\
 &  & SID & 86.90 & 86.00 & 92.34 & 80.47 & 6.67 & 43.57 \\
 &  & \modelname & 89.70 & 89.12 & 94.44 & 84.37 & 4.97 & 44.67 \\
\cmidrule(l){2-9}
 & \multirow{8}{*}{Adversarial} & Baseline & 83.37 & 81.57 & 91.47 & 73.60 & 6.86 & 40.23 \\
 &  & DoLa & 83.47 & 81.51 & 92.47 & 72.87 & 5.93 & 39.40 \\
 &  & OPERA & 84.78 & 83.45 & 91.44 & 76.74 & 7.18 & 41.96 \\
 &  & VCD & 84.26 & 83.90 & 85.86 & 82.02 & 13.50 & 47.76 \\
 &  & ICT & 86.16 & 84.32 & 97.25 & 74.43 & 2.11 & 38.27 \\
 &  & MaskCD & 86.05 & 86.15 & 85.54 & 86.77 & 14.67 & 50.72 \\
 &  & SID & 83.83 & 83.03 & 87.35 & 79.12 & 11.46 & 45.29 \\
 &  & \modelname & 87.83 & 87.39 & 90.67 & 84.34 & 8.68 & 46.51 \\
\midrule
\multirow{24}{*}{A-OKVQA} & \multirow{8}{*}{Random} & Baseline & 86.40 & 85.07 & 94.29 & 77.49 & 4.69 & 41.09 \\
 &  & DoLa & 87.40 & 86.46 & 93.43 & 80.46 & 5.66 & 43.06 \\
 &  & OPERA & 88.19 & 88.43 & 86.67 & 90.26 & 13.88 & 52.07 \\
 &  & VCD & 89.22 & 89.01 & 90.78 & 87.31 & 8.87 & 48.09 \\
 &  & ICT & 88.96 & 87.90 & 97.24 & 80.20 & 2.28 & 41.24 \\
 &  & MaskCD & 90.55 & 90.60 & 90.12 & 91.08 & 9.98 & 50.53 \\
 &  & SID & 88.49 & 87.93 & 92.43 & 83.85 & 6.87 & 45.36 \\
 &  & \modelname & 92.10 & 91.93 & 93.95 & 89.99 & 5.79 & 47.89 \\
\cmidrule(l){2-9}
 & \multirow{8}{*}{Popular} & Baseline & 85.77 & 84.49 & 92.84 & 77.52 & 5.98 & 41.75 \\
 &  & DoLa & 87.53 & 86.06 & 97.56 & 76.98 & 1.92 & 39.45 \\
 &  & OPERA & 87.91 & 87.13 & 93.14 & 81.85 & 6.03 & 43.94 \\
 &  & VCD & 87.85 & 87.81 & 88.10 & 87.52 & 11.82 & 49.67 \\
 &  & ICT & 87.43 & 86.39 & 94.18 & 79.79 & 4.93 & 42.36 \\
 &  & MaskCD & 89.05 & 89.15 & 88.34 & 89.97 & 11.87 & 50.92 \\
 &  & SID & 87.62 & 87.41 & 88.92 & 85.95 & 10.71 & 48.33 \\
 &  & \modelname & 89.67 & 89.70 & 89.44 & 89.96 & 10.62 & 50.29 \\
\cmidrule(l){2-9}
 & \multirow{8}{*}{Adversarial} & Baseline & 80.37 & 79.68 & 82.58 & 76.97 & 16.23 & 46.60 \\
 &  & DoLa & 81.83 & 81.33 & 83.63 & 79.15 & 15.49 & 47.32 \\
 &  & OPERA & 80.82 & 81.54 & 78.59 & 84.72 & 23.08 & 53.90 \\
 &  & VCD & 81.27 & 82.38 & 77.77 & 87.57 & 25.03 & 56.30 \\
 &  & ICT & 83.60 & 83.02 & 86.06 & 80.18 & 12.98 & 46.58 \\
 &  & MaskCD & 82.75 & 83.95 & 78.49 & 90.23 & 24.73 & 57.48 \\
 &  & SID & 80.84 & 81.51 & 78.76 & 84.46 & 22.78 & 53.62 \\
 &  & \modelname & 83.63 & 84.61 & 79.83 & 90.00 & 22.74 & 56.37 \\
\midrule
\multirow{24}{*}{GQA} & \multirow{8}{*}{Random} & Baseline & 85.10 & 83.87 & 91.42 & 77.47 & 7.27 & 42.37 \\
 &  & DoLa & 87.13 & 86.05 & 93.93 & 79.39 & 5.13 & 42.26 \\
 &  & OPERA & 86.02 & 85.29 & 89.99 & 81.06 & 9.02 & 45.04 \\
 &  & VCD & 85.59 & 85.33 & 86.90 & 83.82 & 12.64 & 48.23 \\
 &  & ICT & 88.96 & 87.89 & 97.32 & 80.12 & 2.20 & 41.16 \\
 &  & MaskCD & 89.25 & 89.10 & 90.36 & 87.87 & 9.37 & 48.62 \\
 &  & SID & 84.86 & 84.25 & 87.79 & 80.99 & 11.27 & 46.13 \\
 &  & \modelname & 92.23 & 92.09 & 93.78 & 90.46 & 6.00 & 48.23 \\
\cmidrule(l){2-9}
 & \multirow{8}{*}{Popular} & Baseline & 80.87 & 80.33 & 82.66 & 78.12 & 16.38 & 47.25 \\
 &  & DoLa & 82.53 & 82.35 & 83.21 & 81.51 & 16.45 & 48.98 \\
 &  & OPERA & 81.97 & 82.12 & 81.44 & 82.81 & 18.87 & 50.84 \\
 &  & VCD & 81.83 & 82.23 & 80.46 & 84.08 & 20.42 & 52.25 \\
 &  & ICT & 86.43 & 85.47 & 91.98 & 79.82 & 6.96 & 43.39 \\
 &  & MaskCD & 86.35 & 86.70 & 84.53 & 88.98 & 16.28 & 52.63 \\
 &  & SID & 81.61 & 81.83 & 80.86 & 82.82 & 19.60 & 51.21 \\
 &  & \modelname & 88.37 & 88.60 & 86.88 & 90.39 & 13.65 & 52.02 \\
\cmidrule(l){2-9}
 & \multirow{8}{*}{Adversarial} & Baseline & 78.77 & 78.56 & 79.34 & 77.79 & 20.25 & 49.02 \\
 &  & DoLa & 82.00 & 81.51 & 83.79 & 79.35 & 15.35 & 47.35 \\
 &  & OPERA & 80.24 & 80.64 & 79.04 & 82.31 & 21.83 & 52.07 \\
 &  & VCD & 80.01 & 80.75 & 77.87 & 83.85 & 23.83 & 53.84 \\
 &  & ICT & 84.10 & 83.53 & 86.64 & 80.64 & 12.44 & 46.54 \\
 &  & MaskCD & 83.25 & 84.30 & 79.33 & 89.94 & 23.44 & 56.69 \\
 &  & SID & 79.58 & 79.88 & 78.72 & 81.07 & 21.91 & 51.49 \\
 &  & \modelname & 84.27 & 85.18 & 80.52 & 90.41 & 21.87 & 56.14 \\
\bottomrule
\end{tabular*}
\caption{POPE decomposition on Qwen2.5-VL. \modelname's Yes-rate rises on every split, but accuracy rises jointly even on Yes-biased splits where a pure Yes/No rebalancing would predict accuracy loss.}
\label{tab:pope-decomp-qwen25vl}
\end{table}

\begin{table}[!ht]
\centering
\scriptsize
\setlength{\tabcolsep}{4.0pt}
\renewcommand{\arraystretch}{0.92}
\begin{tabular*}{\textwidth}{@{\extracolsep{\fill}}lllrrrrrr@{}}
\toprule
Dataset & Setting & Method & Acc & F1 & Prec.\,\% & Recall\,\% & FPR\,\% & Yes\,\% \\
\midrule
\multirow{24}{*}{COCO} & \multirow{8}{*}{Random} & Baseline & 83.29 & 81.33 & 92.14 & 72.79 & 6.21 & 39.50 \\
 &  & DoLa & 85.97 & 86.14 & 85.11 & 87.20 & 15.26 & 51.23 \\
 &  & OPERA & 89.20 & 88.81 & 92.14 & 85.71 & 7.31 & 46.51 \\
 &  & VCD & 87.73 & 87.16 & 91.41 & 83.29 & 7.83 & 45.56 \\
 &  & ICT & 89.18 & 88.48 & 94.60 & 83.10 & 4.74 & 43.92 \\
 &  & MaskCD & 88.55 & 88.30 & 90.27 & 86.41 & 9.31 & 47.86 \\
 &  & SID & 88.05 & 87.34 & 92.86 & 82.44 & 6.34 & 44.39 \\
 &  & \modelname & 89.27 & 89.18 & 89.93 & 88.44 & 9.90 & 49.17 \\
\cmidrule(l){2-9}
 & \multirow{8}{*}{Popular} & Baseline & 81.88 & 80.06 & 89.00 & 72.75 & 8.99 & 40.87 \\
 &  & DoLa & 82.93 & 83.80 & 79.74 & 88.30 & 22.44 & 55.37 \\
 &  & OPERA & 86.64 & 84.89 & 97.69 & 75.06 & 1.78 & 38.42 \\
 &  & VCD & 85.38 & 85.06 & 86.96 & 83.24 & 12.48 & 47.86 \\
 &  & ICT & 86.07 & 86.40 & 84.40 & 88.50 & 16.36 & 52.43 \\
 &  & MaskCD & 86.25 & 86.00 & 87.59 & 84.46 & 11.96 & 48.21 \\
 &  & SID & 85.83 & 85.44 & 87.86 & 83.15 & 11.49 & 47.32 \\
 &  & \modelname & 86.83 & 86.51 & 88.66 & 84.46 & 10.80 & 47.63 \\
\cmidrule(l){2-9}
 & \multirow{8}{*}{Adversarial} & Baseline & 78.96 & 77.57 & 83.06 & 72.76 & 14.84 & 43.80 \\
 &  & DoLa & 77.17 & 79.41 & 72.31 & 88.05 & 33.71 & 60.88 \\
 &  & OPERA & 81.24 & 81.38 & 80.78 & 81.99 & 19.51 & 50.75 \\
 &  & VCD & 80.88 & 81.33 & 79.46 & 83.29 & 21.53 & 52.41 \\
 &  & ICT & 83.30 & 82.54 & 86.48 & 78.95 & 12.35 & 45.65 \\
 &  & MaskCD & 81.80 & 82.10 & 80.77 & 83.48 & 19.88 & 51.68 \\
 &  & SID & 81.41 & 81.54 & 80.97 & 82.11 & 19.29 & 50.70 \\
 &  & \modelname & 84.07 & 83.41 & 87.02 & 80.09 & 11.95 & 46.02 \\
\midrule
\multirow{24}{*}{A-OKVQA} & \multirow{8}{*}{Random} & Baseline & 83.45 & 82.56 & 87.25 & 78.35 & 11.45 & 44.90 \\
 &  & DoLa & 83.23 & 84.83 & 77.44 & 93.78 & 27.32 & 60.55 \\
 &  & OPERA & 88.03 & 87.00 & 95.19 & 80.11 & 4.05 & 42.08 \\
 &  & VCD & 86.15 & 86.34 & 85.17 & 87.54 & 15.24 & 51.39 \\
 &  & ICT & 89.00 & 88.71 & 91.11 & 86.43 & 8.43 & 47.43 \\
 &  & MaskCD & 87.45 & 87.55 & 86.86 & 88.25 & 13.35 & 50.80 \\
 &  & SID & 86.47 & 86.52 & 86.20 & 86.84 & 13.90 & 50.37 \\
 &  & \modelname & 89.13 & 89.52 & 86.42 & 92.85 & 14.59 & 53.72 \\
\cmidrule(l){2-9}
 & \multirow{8}{*}{Popular} & Baseline & 79.90 & 79.59 & 80.84 & 78.38 & 18.58 & 48.48 \\
 &  & DoLa & 76.47 & 79.86 & 69.80 & 93.30 & 40.36 & 66.83 \\
 &  & OPERA & 83.22 & 84.15 & 79.73 & 89.09 & 22.65 & 55.87 \\
 &  & VCD & 81.85 & 82.82 & 78.62 & 87.50 & 23.80 & 55.65 \\
 &  & ICT & 83.40 & 84.45 & 79.43 & 90.15 & 23.35 & 56.75 \\
 &  & MaskCD & 83.05 & 83.75 & 80.43 & 87.36 & 21.26 & 54.31 \\
 &  & SID & 82.48 & 83.20 & 79.92 & 86.77 & 21.81 & 54.29 \\
 &  & \modelname & 83.77 & 84.73 & 80.00 & 90.06 & 22.52 & 56.29 \\
\cmidrule(l){2-9}
 & \multirow{8}{*}{Adversarial} & Baseline & 74.04 & 75.15 & 72.07 & 78.51 & 30.43 & 54.47 \\
 &  & DoLa & 68.03 & 74.49 & 61.97 & 93.35 & 57.29 & 75.32 \\
 &  & OPERA & 73.82 & 77.91 & 67.38 & 92.34 & 44.70 & 68.52 \\
 &  & VCD & 74.97 & 77.73 & 70.01 & 87.36 & 37.42 & 62.39 \\
 &  & ICT & 75.56 & 78.68 & 69.77 & 90.19 & 39.07 & 64.63 \\
 &  & MaskCD & 75.90 & 78.40 & 71.03 & 87.47 & 35.67 & 61.57 \\
 &  & SID & 75.50 & 77.94 & 70.88 & 86.56 & 35.56 & 61.06 \\
 &  & \modelname & 77.30 & 79.43 & 72.62 & 87.65 & 33.05 & 60.35 \\
\midrule
\multirow{24}{*}{GQA} & \multirow{8}{*}{Random} & Baseline & 83.73 & 82.95 & 87.13 & 79.16 & 11.70 & 45.43 \\
 &  & DoLa & 83.70 & 85.29 & 77.71 & 94.51 & 27.11 & 60.81 \\
 &  & OPERA & 88.13 & 88.91 & 83.43 & 95.16 & 18.90 & 57.03 \\
 &  & VCD & 86.65 & 86.99 & 84.83 & 89.26 & 15.96 & 52.61 \\
 &  & ICT & 89.23 & 89.34 & 88.44 & 90.26 & 11.80 & 51.03 \\
 &  & MaskCD & 87.85 & 88.25 & 85.44 & 91.25 & 15.55 & 53.40 \\
 &  & SID & 86.97 & 87.17 & 85.85 & 88.53 & 14.59 & 51.56 \\
 &  & \modelname & 89.47 & 89.53 & 89.02 & 90.04 & 11.10 & 50.57 \\
\cmidrule(l){2-9}
 & \multirow{8}{*}{Popular} & Baseline & 78.17 & 78.37 & 77.66 & 79.09 & 22.75 & 50.92 \\
 &  & DoLa & 74.03 & 78.75 & 66.64 & 96.24 & 48.18 & 72.21 \\
 &  & OPERA & 79.27 & 82.11 & 72.22 & 95.14 & 36.60 & 65.87 \\
 &  & VCD & 80.73 & 82.24 & 76.26 & 89.23 & 27.77 & 58.50 \\
 &  & ICT & 80.86 & 82.64 & 75.61 & 91.11 & 29.39 & 60.25 \\
 &  & MaskCD & 82.05 & 83.10 & 78.51 & 88.26 & 24.16 & 56.21 \\
 &  & SID & 81.36 & 82.62 & 77.39 & 88.61 & 25.89 & 57.25 \\
 &  & \modelname & 82.93 & 83.75 & 79.91 & 87.98 & 22.12 & 55.05 \\
\cmidrule(l){2-9}
 & \multirow{8}{*}{Adversarial} & Baseline & 75.08 & 76.06 & 73.18 & 79.17 & 29.01 & 54.09 \\
 &  & DoLa & 68.73 & 74.78 & 62.66 & 92.72 & 55.26 & 73.99 \\
 &  & OPERA & 75.00 & 78.71 & 68.54 & 92.43 & 42.43 & 67.43 \\
 &  & VCD & 76.09 & 78.78 & 70.81 & 88.77 & 36.59 & 62.68 \\
 &  & ICT & 77.40 & 80.11 & 71.53 & 91.02 & 36.22 & 63.62 \\
 &  & MaskCD & 77.10 & 79.30 & 72.35 & 87.73 & 33.53 & 60.63 \\
 &  & SID & 76.62 & 78.99 & 71.72 & 87.90 & 34.66 & 61.28 \\
 &  & \modelname & 79.60 & 81.17 & 75.37 & 87.94 & 28.74 & 58.34 \\
\bottomrule
\end{tabular*}
\caption{POPE decomposition on LLaVA-v1.5. \modelname's Yes-rate rises on every split, but accuracy rises jointly even on Yes-biased splits where a pure Yes/No rebalancing would predict accuracy loss.}
\label{tab:pope-decomp-llavav15}
\end{table}

\clearpage
\section{Additional Case Study}
\label{app:cases}

\begin{figure}[!htbp]
\centering
\includegraphics[width=\linewidth]{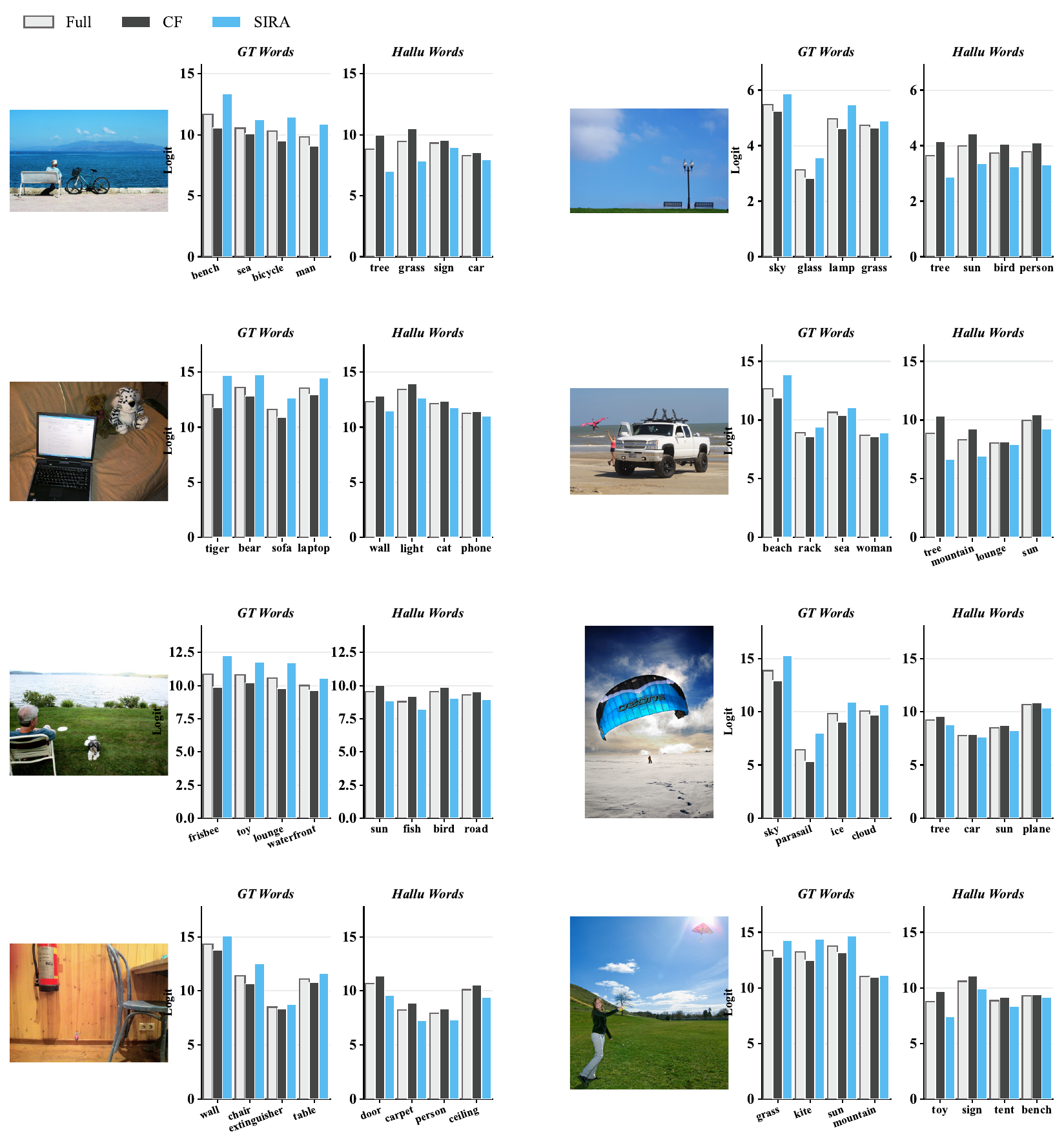}
\caption{Additional case studies on the AMBER benchmark. Eight AMBER cases are shown; for each case we display the input image, the GT-word bar chart, and the Hallu-word bar chart, comparing the Full, CF, and \modelname logits at the token level.}
\label{fig:case-study-appendix}
\end{figure}

We provide additional logit-level case studies on the AMBER benchmark in Figure~\ref{fig:case-study-appendix}. Eight cases generated by Qwen2.5-VL are shown under the same caption-mode setting. For each case we display the input image and two logit-delta bar charts: one for ground-truth object tokens (GT Words) and one for the hallucinated tokens emitted by the unmodified backbone (Hallu Words); within each chart the Full and CF bars report the base model's logit under the full multimodal context and under the post-boundary image-masked counterfactual, and the \modelname bar reports the contrastive logit $(1{+}\alpha)\,z^{\text{full}} - \alpha\,z^{\text{cf}}$ actually used for decoding.

Three patterns recur across all eight cases and are worth calling out explicitly. (i)~On GT Words, the Full bar consistently exceeds CF by a clear margin and \modelname pushes the value further up, so the contrast \emph{reinforces} visually grounded predictions rather than flipping them. (ii)~On Hallu Words, CF is comparable to or strictly higher than Full on a non-trivial fraction of tokens---the cleanest token-level signature of language-prior-driven hallucination, since these tokens remain probable even after the image is masked away. \modelname collapses such tokens to the lowest of the three bars. (iii)~On GT Words where Full $\approx$ CF, the \modelname bar barely moves; the contrastive correction is therefore self-gating, leaving already image-independent-but-correct tokens untouched and concentrating its effect on the tokens that actually need it.

These token-level observations are the microscopic counterpart of the POPE Yes-rate decomposition in Appendix~\ref{app:pope-decomp}: the \emph{simultaneous} rise of Yes-rate and accuracy on Yes-biased POPE splits is not a uniform Yes/No rebalancing but the aggregate effect of pattern~(ii) suppressing prior-driven false positives while pattern~(i) preserves true positives. Figure~\ref{fig:case-study-appendix} thus links the macroscopic behavior quantified in Section~\ref{sec:exp-pope} and Table~\ref{tab:pope-decomp-qwen25vl} to a concrete token-level mechanism.

\end{document}